\pdfoutput=1

\documentclass[11pt]{article}

\usepackage[final]{acl}

\usepackage{times}
\usepackage{latexsym}

\usepackage[T1]{fontenc}

\usepackage[utf8]{inputenc}

\usepackage{microtype}

\usepackage{inconsolata}

\usepackage{graphicx}

\title{Skeletons Matter: Dynamic Data Augmentation for Text-to-Query}

\author{
 \textbf{Yuchen Ji\textsuperscript{1$\heartsuit$}},
 \textbf{Bo Xu\textsuperscript{2,5}}\thanks{Corresponding authors.},
 \textbf{Jie Shi\textsuperscript{3,5$\heartsuit$}},
 \textbf{Jiaqing Liang\textsuperscript{1,5$\spadesuit$}}\footnotemark[1],
 \textbf{Deqing Yang\textsuperscript{1,5$\spadesuit$}}, 
 \textbf{Yu Mao\textsuperscript{4}}, \\
 \textbf{Hai Chen\textsuperscript{4}},
 \textbf{Yanghua Xiao\textsuperscript{3,5$\spadesuit$}}
\\
 \textsuperscript{1}School of Data Science, Fudan University \\
\textsuperscript{2}School of Computer Science and Technology, Donghua University \\
 \textsuperscript{3}College of Computer Science and Artificial Intelligence, Fudan University \\
 \textsuperscript{4}Ant Group, 
 \textsuperscript{5}Shanghai Key Laboratory of Data Science\\
 \texttt{\textsuperscript{$\heartsuit$}\{ycji24,jshi22\}@m.fudan.edu.cn}, \texttt{\textsuperscript{2}xubo@dhu.edu.cn} \\
 \texttt{\textsuperscript{$\spadesuit$}\{liangjiaqing,yangdeqing,shawyh\}@fudan.edu.cn}\\ \texttt{\textsuperscript{4}\{songhao.my,chenhai.ch\}@antgroup.com}
}

\usepackage{multirow}
\usepackage[table,xcdraw]{xcolor}
\usepackage[normalem]{ulem}
\useunder{\uline}{\ul}{}
\usepackage{array}
\usepackage{colortbl}
\usepackage{booktabs}

\begin{document}
\maketitle
\begin{abstract}
The task of translating natural language questions into query languages has long been a central focus in semantic parsing. Recent advancements in Large Language Models (LLMs) have significantly accelerated progress in this field. However, existing studies typically focus on a single query language, resulting in methods with limited generalizability across different languages. In this paper, we formally define the Text-to-Query task paradigm, unifying semantic parsing tasks across various query languages. We identify query skeletons as a shared optimization target of Text-to-Query tasks, and propose a general dynamic data augmentation framework that explicitly diagnoses model-specific weaknesses in handling these skeletons to synthesize targeted training data. Experiments on four Text-to-Query benchmarks demonstrate that our method achieves state-of-the-art performance using only a small amount of synthesized data, highlighting the efficiency and generality of our approach and laying a solid foundation for unified research on Text-to-Query tasks. We release our code at \url{https://github.com/jjjycaptain/Skeletron}
\end{abstract}

\section{Introduction}
\begin{figure}[t]
  \includegraphics[width=\columnwidth]{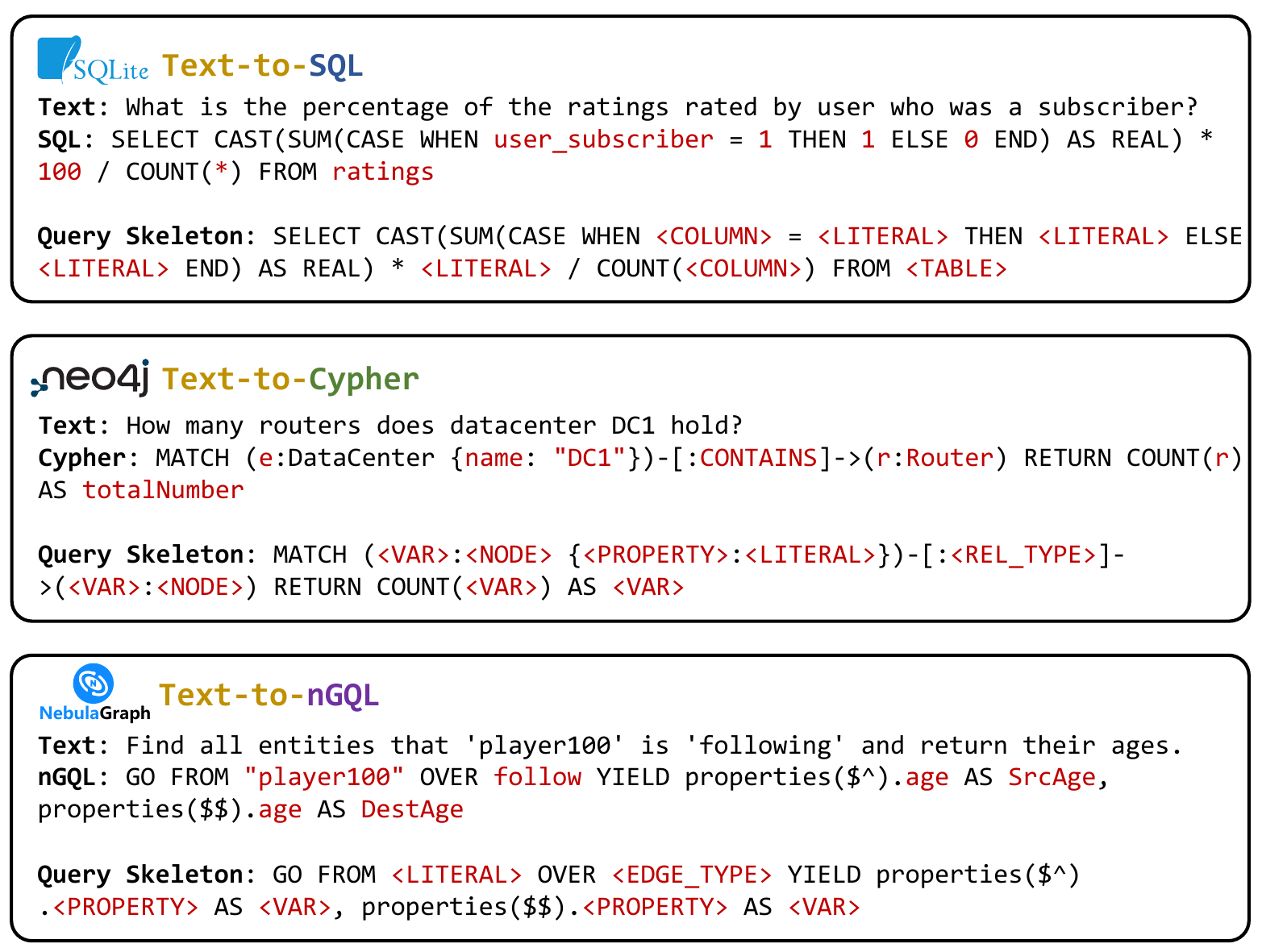}
  \caption{Examples of query skeletons from three different query languages.}
  \label{fig:fsk-intro}
\end{figure}

The task of translating natural language questions into query languages (e.g., Text-to-SQL, Text-to-Cypher, Text-to-nGQL) has long been a central focus in semantic parsing \citep{popescu-etal-2004-modern,wikidata,spccql,nl2gql}. It aims to facilitate user interaction with databases by allowing input in natural language, thereby improving the efficiency of data access. Given this shared objective and task formulation, in this paper, we unify these related tasks under a single task paradigm, \textbf{Text-to-Query}, and develop general methods for this unified setting. This broader perspective invites us to examine common challenges and optimization opportunities in formal query generation across different query languages. While concrete queries are tied to specific schemas and databases, many of them share the same underlying syntactic and semantic structures once instance-specific elements are stripped away. These abstract structures, which we refer to as \textbf{query skeletons}, reveal recurring patterns in how queries are composed across diverse contexts (see in Figure \ref{fig:fsk-intro}). We argue that query skeletons serve as a key abstraction for understanding model behavior, diagnosing failure cases, and designing generalizable optimization strategies for Text-to-Query tasks.

Recently, the advancement of Large Language Models (LLMs; \citealp{openai2024gpt4technicalreport,qwen2025qwen25technicalreport}) has significantly accelerated progress in Text-to-Query tasks. Current approaches can be broadly categorized into In-Context Learning (ICL) and Fine-Tuning (FT) paradigms. Specifically, ICL-based methods \citep{pourreza2023dinsql,dail_sql,wang-etal-2024-improving-demonstration} rely on sophisticated prompt engineering to guide proprietary LLMs in generating queries, achieving impressive accuracy. However, these methods face concerns regarding data privacy and high inference costs. As an alternative, FT-based methods leverage open-source LLMs and improve their performance through incremental pre-training and supervised fine-tuning (SFT) \citep{codes,pourreza-rafiei-2024-dts}. Many of these methods construct training data using LLM-based data augmentation \citep{sense,li2025omnisqlsynthesizinghighqualitytexttosql,tiwari-etal-2025-auto,zhong-etal-2025-synthet2c}. 

Despite their promise, existing data augmentation approaches suffer from several key limitations: (i) They overlook the critical value of query skeletons; (ii) Their strategies are static, lacking adaptation to different target model needs, which may result in redundant data with limited benefit, often sampling question types the model already handles well; (iii) They primarily focus on a single query language, which may hinder their applicability to other Text-to-Query tasks. To address these limitations, we propose a \textbf{dynamic data augmentation method based on query skeletons for Text-to-Query tasks.}

Inspired by the theory of diagnostic teaching \citep{reynolds2007diagnostic} in educational psychology, our approach begins with dynamically diagnosing the weaknesses of a target LLM in a target dataset. We first diagnose model failures on the training set to identify query skeletons it struggles with, forming an error-prone skeleton set that reveals its systematic weaknesses. Additionally, To avoid overfitting the synthesized data to a narrow set of skeletons, we train a skeleton generator on the error-prone set to produce novel ones, expanding the set into a more diverse candidate skeleton pool. Then, we introduce a skeleton-guided backward-forward data synthesis pipeline, where concrete queries are instantiated from skeletons and back-translated into natural language questions, then verified by reasoning forward from the questions to ensure consistency using chain-of-thought (CoT; \citealp{cot1,cot2}) prompting. Finally, the data synthesized through the pipeline are used to fine-tune the target LLM, thereby enhancing its understanding of the previously misaligned query skeletons.

In summary, our contributions are threefold:

\begin{itemize}
    \item We are the first to formally define and systematize the Text-to-Query task paradigm, unifying semantic parsing across a broad range of query languages, and laying the foundation for unified method development.

    \item We propose a unified data augmentation framework for Text-to-Query tasks that dynamically identifies the query skeletons a model struggles with and generates targeted training examples accordingly, enabling both behavioral analysis and performance improvement across query languages.

    \item Our method achieves state-of-the-art performance on Four Text-to-Query benchmarks (Spider, BIRD, Text2Cypher and NL2GQL), demonstrating its effectiveness and generality across different Text-to-Query tasks.
    
\end{itemize}

\section{Related Work}
\paragraph{Text-to-Query Based On LLM} 
Currently, many Text-to-Query methods are built on the powerful reasoning ability of LLMs. A significant portion of these methods rely on ICL. Some studies select few-shot examples based on input similarity to guide inference \citep{fewshot1,fewshot2,fewshot3}, while others reduce task complexity by decomposing tasks or questions into simpler substeps \citep{pourreza2023dinsql,decomp1,decomp2,decomp3}. Additional works enhance reasoning capabilities through strategies like CoT \citep{cotq1,cotq2} and consistency-driven reasoning \citep{decomp1,dail_sql}. However, these ICL methods typically rely on proprietary LLMs, raising concerns about privacy risks and inference costs. To enhance open-source models' Text-to-Query abilities, CODES \citep{codes} proposed incremental pretraining on hybrid corpus. Nevertheless, incremental pretraining is resource-intensive and collecting sufficient training corpus is challenging for SQL and even harder for specialized query languages, limiting its applicability across diverse Text-to-Query tasks.

\paragraph{Text-to-Query Data Augmentation} 
High-quality Text-to-Query datasets remain scarce due to the high cost of manual annotation. To mitigate this, many approaches adopt data augmentation to automatically generate examples. Early methods synthesize queries using context-free grammars (CFGs) or rule-based slot filling over SQL skeletons, followed by back translation into natural language questions using Pretrained Language Models (PLMs) or seq2seq models \citep{back-trans1,back-trans2,cfg2,zhong-etal-2020-grounded}. However, these approaches rely on manually crafted CFGs and language-specific rules, limiting their generalizability across Text-to-Query tasks. Moreover, the limitations of conventional neural models often lead to unnatural questions. Recent work typically employs LLMs to synthesize data. SENSE \citep{sense}, OmniSQL \citep{li2025omnisqlsynthesizinghighqualitytexttosql}, Auto-Cypher \citep{tiwari-etal-2025-auto}, and SyntheT2C \citep{zhong-etal-2025-synthet2c} design elaborate pipelines based on LLMs to synthesize high-quality data. Compared with our method, these methods lack explicit modeling and utilization of the query skeleton and follow static generation strategies, which leads to redundancy and limited benefit of augmented data.

\section{Task Formulation}
To support theoretical modeling and general-purpose solution development, We formally define the \textbf{Text-to-Query} task as:
\[
f(S, q) \rightarrow Q
\]
where \( q \) is the input question, \( S \) is the database schema, and \( Q \) is the generated query in a language such as SQL, Cypher, or nGQL. The schema \( S \) provides structural and semantic context necessary for interpreting the question. Its representation depends on the underlying data model. The following are illustrative examples of schema formulations for common database types:

For relational databases (e.g., SQLite), the schema can be represented as \( S = \{(t, c, \tau) \mid t \in \mathcal{T},\ c \in \mathcal{C}_t,\ \tau \in \mathcal{D} \} \), where \( \mathcal{T} \) denotes the set of table names, \( \mathcal{C}_t \) is the set of columns in table \( t \), and \( \tau \) is the data type of column \( c \).

For graph-based databases (e.g., Neo4j or NebulaGraph), the schema can be represented as \( S = \{(e_1, r, e_2) \mid e_1, e_2 \in \mathcal{E},\ r \in \mathcal{R} \} \), where \( \mathcal{E} \) is the set of node types, and \( \mathcal{R} \) is the set of relation types.

This formulation provides a unified foundation for developing Text-to-Query models across heterogeneous query languages and databases.

\begin{figure*}[t]
\centering
\includegraphics[width=1\linewidth] {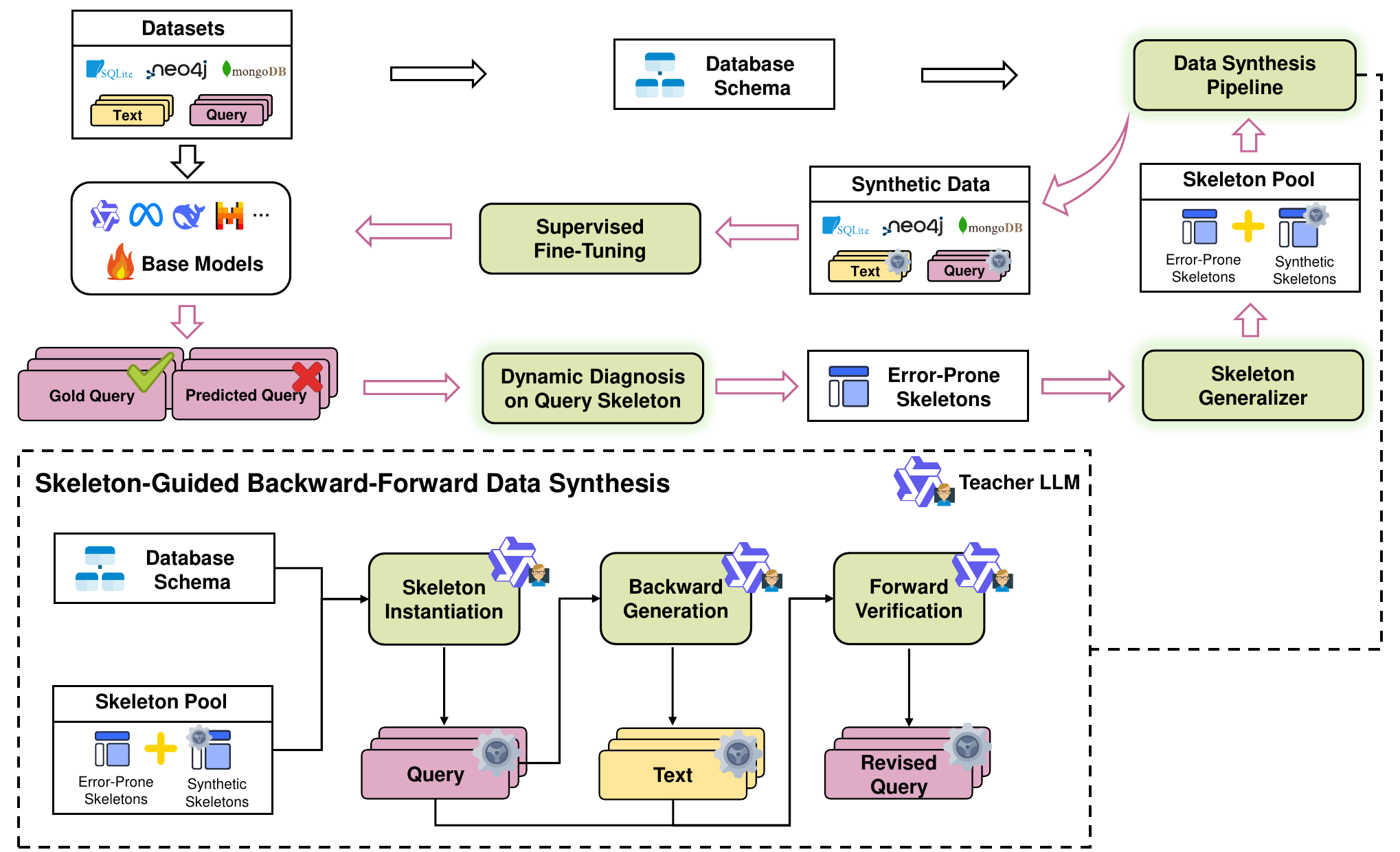}
\caption{Our proposed method consists of three key components: 
(i) \textbf{Dynamic Diagnosis on Query Skeletons}: We analyze model behavior to identify query skeletons it struggles with, constructing an error-prone skeleton set to guide targeted data synthesis.
(ii)\textbf{Skeleton Generalizer:} A skeleton generation model is trained on the error-prone set to produce structurally novel skeletons, expanding the diversity of the skeleton pool.
(iii) \textbf{Skeleton-Guided Backward-Forward Data Synthesis}: We instantiate skeletons from the pool under diverse schema contexts and synthesize high-quality, targeted training data through a backward-forward generation framework.}
  \label{fig:overview}
\end{figure*}

\section{Method}
An overview of our proposed dynamic data augmentation method based on formal query skeletons is shown in Figure \ref{fig:overview}.

\subsection{Dynamic Diagnosis on Query Skeletons}
\label{sec:error-detect}
Unlike existing Text-to-Query data augmentation methods, our approach is dynamic: it introduces a diagnostic step before augmentation to identify model-specific weaknesses, enabling more targeted, intelligent, and efficient data synthesis.

Given a target LLM and a Text-to-Query dataset, we first perform $K$-fold cross-validation on the training set to identify cases where the model fails. However, these failure cases can arise from a wide range of issues, including schema-linking errors, misunderstandings of database content, and syntactic mistakes, as noted in prior work \citep{liu2025surveynl2sqllargelanguage,bird}. Since our goal is to diagnose the ability of an LLM to handle query skeletons, we aim to isolate and focus specifically on this type of error during the diagnostic process.

To achieve this goal, we introduce a \textbf{structural similarity measure} to detect whether a model has generated the correct query skeleton. Specifically, we provide two implementations of this measure, \textbf{AST-based structural distance} and \textbf{Token-based structural distance}, depending on the availability of parsing tools for the target query language.

\paragraph{AST-Based Structural Distance} Abstract syntax trees (ASTs) represent the hierarchical structure of code in tree form and are widely used in program analysis for measuring code similarity \citep{codeast1,codeast2}. They also serve as a common intermediate representation for parsing query languages \citep{sciencebenchmark,sqlast}. In this setting, we parse both the predicted and gold queries into ASTs and compare their structural differences. Specifically, we apply the Change Distiller algorithm \citep{changedistll} to compute the minimum set of edit operations (e.g., insert, delete, update, keep, etc.) required to transform one AST into another. We define the AST-based structural distance as the total number of non-keep operations, which reflects the degree of structural discrepancy between the two query skeletons.

\paragraph{Token-Based Structural Distance} In principle, all query languages can be parsed into ASTs, as their syntax is inherently hierarchical. However, some less commonly used query languages (e.g., nGQL) lack a mature ecosystem, and open-source parsers for these languages are often unavailable. This creates practical engineering barriers to implementing the AST-based structural distance, even though the method itself remains theoretically applicable. In such scenarios, we provide a structural similarity measure that compares the predicted and gold skeletons using token-level edit distance. Although this approximation is less fine-grained, it still captures structural divergence to a reasonable extent and enables the diagnostic framework to remain applicable across a wide range of query languages. Implementation details of the two structural similarity measures are presented in the Appendix~\ref{app:measure}.

\paragraph{Skeleton Error Detection Based on Structural Similarity Measurement} The most straightforward way to detect query skeleton errors is to check whether the predicted and gold skeletons yield identical structures. However, this criterion is overly strict and may lead to false positives. Through our analysis of prediction errors, we observed that some predicted and gold skeletons differ slightly in structure but remain semantically equivalent—for example, differing only in the presence of a \texttt{DISTINCT} keyword or a change in a single operator. To mitigate such cases, we introduce a relaxed threshold-based criterion: if the structural distance exceeds a threshold, we classify the sample as a query skeleton error. More discussion on threshold selection is provided in Section~\ref{sec:analysis}.

Finally, we select skeletons with an error rate above 20\% to construct the error-prone skeleton set, which serves as the foundation for the subsequent construction of the skeleton generalizer and data synthesis.

\subsection{Skeleton Generalizer}
Although the error-prone skeleton set already contains a rich and realistic collection of query skeletons, it is inevitable that novel skeletons will appear in test scenarios. If data augmentation is performed using only error-prone skeleton set, the resulting model may fail to handle unseen patterns during evaluation. To address this limitation, We propose to use a skeleton generalizer to generate novel but structurally meaningful skeletons that go beyond the error-prone set.

Specifically, we fine-tune an LLM using the previously collected error-prone skeleton set to learn their underlying patterns, thereby constructing a skeleton generation model capable of producing new skeletons. Building on prior work \citep{xu2024wizardlm,scalequest}, we extract a portion of the LLM’s instruction template (e.g., “\texttt{<im\_start>Assistant:}”) as a prefix to guide 
skeleton generation, and fine-tune the model on $(prefix, skeleton)$ pairs constructed from the error-prone skeleton set. During inference, we follow the same prompt format to induce the generation of novel skeletons. These generated skeletons are then combined with the original error-prone ones to form a comprehensive skeleton pool for data synthesis. Further details about the Skeleton Generalizer can be found in the Appendix~\ref{app:generalizer}.

\subsection{Skeleton-Guided Backward-Forward Data Synthesis}
To leverage the query skeletons and dynamic diagnosis results, we perform controlled data synthesis with a teacher LLM guided by the constructed skeleton pool. A backward-forward generation framework is adopted to ensure data quality and reliability. The synthesis process consists of the following key steps:

\paragraph{Skeleton Instantiation} For each database in the target dataset, we randomly sample a query skeleton from the skeleton pool and prompt the teacher LLM to instantiate it by filling in appropriate schema elements (e.g., tables, columns and nodes) from given database schema. Once the query instantiation is complete, we apply rule-based verification to identify and filter out basic errors such as syntax mistakes, execution failures, and invalid join conditions. This process includes verifying query executability, and checking whether the referenced tables and columns satisfy necessary foreign key constraints (for the Text-to-SQL task).

\paragraph{Backward Generation} In this phase, the teacher LLM is prompted to translate the completed query into a corresponding natural language question according to the database schema. Since query languages are formal and semantically unambiguous, this backward translation is substantially easier than the forward direction (i.e., generating queries from natural language), which requires resolving ambiguity in user intent and performing complex schema linking. The clarity of query languages and the relative simplicity of backward generation help ensure the quality of the synthesized question-query pairs.

\paragraph{Forward Verification} Although the skeleton instantiation and backward generation steps provide a reasonable degree of quality assurance, large LLMs can still suffer from hallucinations \citep{halluc1,halluc2}, which may lead to mismatches between the synthesized questions and their corresponding queries. To mitigate this issue, we introduce a forward verification phase, where the teacher LLM is prompted to assess the semantic consistency between the synthesized question and queries using chain-of-thought reasoning \citep{cot1,cot2}, and revise the query if necessary. This process enhances the reliability of the final synthetic dataset.

Finally, we select two open-source models, Qwen2.5-Coder-7B and Qwen2.5-Coder-14B, as base models, and perform SFT using training data synthesized by our data augmentation method. The input sequence for SFT consists of the task description, database schema, and question. We refer to the resulting series of Text-to-Query LLMs as \textbf{Skeletron}. The prompts used in the data synthesis stage are provided in the Appendix.

\section{Experiments}
\subsection{Evaluation Benchmarks}
We evaluate our method on three representative Text-to-Query tasks: Text-to-SQL, Text-to-Cypher and Text-to-nGQL, due to the prominent roles of SQL, Cypher and nGQL in relational and non-relational databases, respectively.

For the Text-to-SQL task, we evaluate our approach on Spider \citep{yu-etal-2018-spider} and BIRD \citep{bird}. Spider is a cross-domain dataset covering 200 databases across 138 domains. BIRD is a more realistic and challenging benchmark, containing 95 databases across 37 professional domains. For Spider, we evaluate on both its development and test sets, while for BIRD, we evaluate only on the development set, as the test set is not publicly available.

For the Text-to-Cypher task, We evaluate our approach on Text2Cypher \citep{ozsoy-etal-2025-text2cypher}, a large-scale dataset released by Neo4j. However, many examples in this dataset lack executable databases, making it difficult to evaluate the correctness of generated queries. As a result, we extract a subset of executable examples to form a new benchmark, Text2Cypher-Exec, which contains 22,093 training samples and 2,471 test samples.

For the Text-to-nGQL task, we evaluate our approach on NL2GQL \citep{nl2gql} dataset. NL2GQL was manually constructed by humans with assistance from LLMs, followed by subsequent refinement to correct errors and enhance naturalness and diversity. The dataset comprises 3,862 training samples and 1,254 test samples.

\subsection{Evaluation Metrics}
For the Text-to-SQL task, following prior work, we use both EX and TS metrics on Spider, and EX metirc on BIRD. EX measures the proportion of predicted SQL queries that produce the same execution results as the corresponding gold queries. TS is a more reliable metric that checks whether a SQL query yields consistent results with the gold query across multiple database variants constructed via data augmentation. Notably, TS is only available on the Spider dev.

For the Text-to-Cypher and Text-to-nGQL task, as no official scripts are available for the corresponding datasets, we compute EX following a similar evaluation procedure as used in BIRD.

\subsection{Baselines}
\label{sec:baselines}
\paragraph{LLMs with Zero-Shot Prompting} We compare our method against both proprietary and open-source LLMs. The proprietary models include GPT-4o, GPT-4-Turbo, and GPT-4o-mini\footnote{Results for GPT-4o, GPT-4-Turbo, and GPT-4o-mini are reported from \citet{li2025omnisqlsynthesizinghighqualitytexttosql}}, while the open-source models include Qwen2 \citep{yang2024qwen2technicalreport}, Qwen2.5 \citep{qwen2025qwen25technicalreport}, Qwen2.5-Coder \citep{hui2024qwen25codertechnicalreport}, and Llama3.3 \citep{grattafiori2024llama3herdmodels}. These models vary in scale and architecture, providing a diverse and representative baseline for evaluation.

\paragraph{FT-Based Methods} We also compare our method with a range of method based on FT. REDSQL \citep{redsql} proposes a method to decouple schema linking and the skeleton parsing. DTS-SQL \citep{pourreza-rafiei-2024-dts} decomposes fine-tuning into schema-linking and SQL generation stages. CODES \citep{codes} employs incremental pre-training along with strategic prompt construction. OmniSQL \citep{li2025omnisqlsynthesizinghighqualitytexttosql} performs SFT using a large-scale dataset of 2.5 million synthetic examples produced by its scalable framework.

\paragraph{Data Augmentation Methods} To conduct a fair comparison with other data augmentation approaches, we adopt the synthetic dataset released by \citet{li2025omnisqlsynthesizinghighqualitytexttosql} and randomly sample a subset of the same size as our synthesized data for SFT. In addition, we construct two static variants of our synthesis pipeline that exclude the dynamic diagnostic step:

\begin{itemize}
    \item Question-to-SQL, which first prompts the LLM to generate a question, then translates it into SQL.
    \item SQL-to-Question, which reverses the order by first generating SQL and then translating it into a corresponding question.
\end{itemize}

All of these methods are evaluated under the same conditions: we apply SFT to the base model using augmented dataset combined with the BIRD original training set without introducing any other optimization techniques, and adopt the same inference settings as used in Skeletron.

\subsection{Implementation Details}
During the dynamic diagnosis, we adopt an AST-based structural similarity measure for the Text-to-SQL task, while using a token-based measure for the Text-to-Cypher and Text-to-nGQL tasks, and set the threshold for skeleton error detection to 2. We use Qwen2.5-Coder-14B-Instruct as the base model to train the skeleton generalizer. For data synthesis, we adopt Qwen2.5-72B-Instruct as the teacher model to generate question-SQL pairs under skeleton constraints. In the fine-tuning stage, we combine the original training set with 10,000 synthesized data and fine-tune the base models with a learning rate of 5e-6, a batch size of 64 and a cosine warmup schedule over 2 epochs. In both the skeleton generator and the final fine-tuning stage, we perform full-parameter fine-tuning using a conditional next-token prediction loss.

During inference, we adopt a zero-shot setting and generate one single prediction per question, using greedy decoding. All experiments are conducted on 8 NVIDIA A800 80GB GPUs.

\begin{table}[t]
\centering
\small 
\begin{tabular}{l|>{\centering\arraybackslash}m{0.65cm}>{\centering\arraybackslash}m{0.65cm}|>{\centering\arraybackslash}m{0.73cm}|>{\centering\arraybackslash}m{0.65cm}}

\toprule
\multicolumn{1}{c|}{} & \multicolumn{2}{c|}{\textbf{Spider Dev}} & \textbf{Spider Test} & \textbf{BIRD Dev} \\ 
\cmidrule(l){2-5} 
\multicolumn{1}{c|}{\multirow{2}{*}[4ex]{\textbf{Model/Method}}} & \textbf{EX} & \textbf{TS} & \textbf{EX} & \textbf{EX} \\ 
\midrule
\multicolumn{5}{c}{\textbf{LLMs (Zero-Shot)}} \\ 
\midrule
GPT-4o-mini & - & 70.4 & 82.4 & 58.8 \\
GPT-4-Turbo & - & 72.4 & 83.4 & 62.0 \\
GPT-4o & - & 70.9 & 83.2 & 61.9 \\
Qwen2.5-72B-Instruct & 83.6 & 74.1 & 85.6 & 58.7 \\
Qwen2-72B-Instruct & 81.5 & 74.2 & 83.3 & 58.5 \\
Llama3.3-70B-Instruct & 77.2 & 68.1 & 75.8 & 60.0 \\
\midrule
\multicolumn{5}{c}{\textbf{FT-Based Methods}} \\
\midrule
RESDSQL-3B & 84.1 & 73.5 & 79.9 & - \\
DTS-SQL 7B & 85.5 & - & 84.4 & 55.8 \\
CODES 7B & 85.4 & 80.3 & - & 57.2 \\
OmniSQL 7B & - & 81.2 & \underline{87.9} & 63.9 \\
CODES 15B & 84.9 & 79.4 & - & 58.5 \\
OmniSQL 14B & - & \underline{81.4} & \textbf{88.3} & \underline{64.2} \\
\midrule
\multicolumn{5}{c}{\textbf{Our Method}} \\
\midrule
\rowcolor[HTML]{EFEFEF} \textbf{Skeletron 7B} & \underline{85.7} & 78.2 & 84.7 & 61.4 \\
\rowcolor[HTML]{EFEFEF} \textbf{Skeletron 14B} & \textbf{87.3} & \textbf{82.0} & 86.6 & \textbf{65.1} \\
\bottomrule
\end{tabular}
\caption{Performance comparison on the Text-to-SQL task. Best results are in \textbf{bold}; second-best are \underline{underlined}.}
\label{tab:main-result}
\end{table}

\begin{table}[t]
\small
\centering
\begin{tabular}{l|cc}
\toprule
\multicolumn{1}{c|}{\multirow{2}{*}{\textbf{Model/Method}}} & \multicolumn{2}{c}{\textbf{EX}}                                  \\ \cmidrule(l){2-3} 
\multicolumn{1}{c|}{}                                       & \multicolumn{1}{c|}{\textbf{Cypher}} & \textbf{nGQL} \\ 
\midrule
\multicolumn{3}{c}{\textbf{LLMs (Zero-Shot)}} \\ 
\midrule
Qwen2.5-72B-Instruct                                        & \multicolumn{1}{c|}{42.9}                      & 26.9            \\
Qwen2-72B-Instruct                                          & \multicolumn{1}{c|}{37.8}                      & 11.1               \\
Llama3.3-70B-Instruct                                       & \multicolumn{1}{c|}{43.3}                      & 18.9                \\
Qwen2.5-Coder-32B-Instruct                                  & \multicolumn{1}{c|}{44.2}                      & 26.5            \\
Qwen2.5-Coder-14B-Instruct                                  & \multicolumn{1}{c|}{39.7}                      & 14.9            \\
Qwen2.5-Coder-7B-Instruct                                   & \multicolumn{1}{c|}{25.9}                      & 5.1             \\
\midrule
\multicolumn{3}{c}{\textbf{Our Method}} \\ 
\midrule
\rowcolor[HTML]{EFEFEF}
\textbf{Skeletron 7B}                                                & \multicolumn{1}{c|}{{\ul 58.4}}                & {\ul 36.7}      \\
\rowcolor[HTML]{EFEFEF}
\textbf{Skeletron 14B}     
& \multicolumn{1}{c|}{\textbf{58.6}}             & \textbf{45.1}                \\ \bottomrule
\end{tabular}
\caption{Performance comparison on additional Text-to-Query tasks. Cypher denotes the Text-to-Cypher task and nGQL denotes the Text-to-nGQL task, where their respective datasets are used as the target datasets for data augmentation.}
\label{tab:cypher}
\end{table}

\begin{table*}[t]
\small
\begin{tabular}{l|*{4}{>{\centering\arraybackslash}p{2cm}}}
\toprule
\multicolumn{1}{c|}{\textbf{Base Model}} & \textbf{simple} & \textbf{moderate} & \textbf{challenging} & \textbf{total} \\ \midrule
\rowcolor[HTML]{EFEFEF} 
Qwen2.5-Coder-7B                   & 45.2            & 22.0              & 20.0                 & 35.8          \\
\quad + BIRD \& Q2S Synthetic Data       & 64.8            & 51.7              & 43.5                 & 58.8           \\
\quad + BIRD \& S2Q Synthetic Data       & 65.7            & 51.9              & 44.8                 & 59.6           \\
\quad+ BIRD \& OmniSQL Synthetic Data   & 60.4            & 41.0              & 32.4                 & 51.9           \\
\quad\textbf{+ BIRD \& Skeletron Synthetic Data} & \textbf{67.6}   & \textbf{53.5}     & \textbf{47.6}        & \textbf{61.4}  \\ \midrule
\rowcolor[HTML]{EFEFEF} 
Qwen2.5-Coder-14B                  & 57.1            & 36.9              & 26.2                 & 48.0           \\
\quad+ BIRD \& Q2S Synthetic Data       & 71.0            & 55.8              & 44.8                 & 64.0           \\
\quad+ BIRD \& S2Q Synthetic Data       & 70.3            & 55.4              & 42.1                 & 63.1           \\
\quad+ BIRD \& OmniSQL Synthetic Data   & 65.4            & 44.4              & 37.9                 & 56.5           \\
\quad\textbf{+ BIRD \& Skeletron Synthetic Data} & \textbf{72.2}   & \textbf{56.0}     & \textbf{49.0}        & \textbf{65.1}  \\ \bottomrule
\end{tabular}
\caption{EX performance of the base model after SFT on data synthesized by different augmentation methods across difficulty levels on the BIRD dev dataset. Q2S and S2Q refer to the Question-to-SQL and SQL-to-Question augmentation strategies described in Section \ref{sec:baselines}, respectively. BIRD denotes the original training data from the BIRD dataset. Each synthetic dataset is limited to 10,000 examples.}
\label{tab:aug}
\end{table*}

\begin{table}[t]
\centering
\scriptsize
\begin{tabular}{l|>{\centering\arraybackslash}m{0.6cm}>
{\centering\arraybackslash}m{0.6cm}|>
{\centering\arraybackslash}m{0.6cm}|>
{\centering\arraybackslash}m{0.6cm}}
\toprule
                         & \multicolumn{2}{c|}{\textbf{Spider Dev}} & \textbf{Spider Test} & \textbf{BIRD Dev} \\ \cmidrule(l){2-5} 
\multirow{-2}{*}{}       & \textbf{EX}         & \textbf{TS}        & \textbf{EX}          & \textbf{EX}       \\ \midrule
\rowcolor[HTML]{EFEFEF} 
\textbf{Skeletron 7B}    & \textbf{85.7}       & \textbf{78.2}      & \textbf{84.7}        & \textbf{61.4}     \\
\quad w/o Synthetic Data       & 83.3                & 75.6               & 82.8                 & 57.5              \\
\quad w/o Dynamic Diagnosis    & 84.3                & 77.7               & 84.1                 & 57.8              \\
\quad w/o Skeleton Generalizer   & 84.3                & 76.5               & 84.7                 & 58.5              \\
\quad w/o Forward Verification & 82.6                & 75.0               & 84.1                 & 59.3              \\ \bottomrule
\end{tabular}
\caption{Ablations on the synthetic data and 3 key components of our method.}
\label{tab:ablation}
\end{table}

\subsection{Main Results}
\paragraph{Results on the Text-to-SQL Task} As shown in Table~\ref{tab:main-result}, Skeletron outperforms all baselines on both Spider and BIRD benchmarks, including its teacher model Qwen2.5-72B-Instruct. Unlike previous FT-based methods that often involve additional optimization techniques such as incremental pre-training or value retrieval, Skeletron achieves comparable or better performance using SFT alone. The only exception is on the Spider test set, where it slightly underperforms OmniSQL. However, OmniSQL uses 2.5 million synthetic examples, while Skeletron uses only 10,000, just 1/250 of the data, yet still surpasses it by +0.6\% TS on Spider dev and +0.9\% EX on BIRD dev.

Table~\ref{tab:aug} presents a fair comparison of data augmentation methods. Our approach yields the largest performance gains across all settings, significantly improving the base model. Under comparable conditions, the gap between Skeletron and OmniSQL widens substantially, reaching up to 9.5\%. It also outperforms both static variants of our method, demonstrating the clear advantage of our synthesis method. Notably, the improvement increases with the difficulty of the question, increasing from 15.1\% to 22.8\%. This benefit comes from the dynamic diagnosis step, which identifies the skeletons the model struggles with (often the more challenging ones) and uses them to construct harder training data.

\paragraph{Results on other Text-to-Query Tasks} As shown in Table~\ref{tab:cypher}, Skeletron 14B also achieves state-of-the-art performance on the Text-to-Cypher and Text-to-nGQL Tasks, surpassing a range of models with significantly larger parameter sizes. In particular, it outperforms the second-best model, Qwen2.5-Coder-32B-Instruct, which is specifically enhanced for code-related tasks and well-suited for query languages, by a margin of 14.4\% on the Text-to-Cypher task and by 18.2\% on the Text-to-nGQL task. These results demonstrate that our method is broadly applicable and effective across the full spectrum of Text-to-Query tasks.

\subsection{Ablation Study}
To assess the contribution of each component, we conduct ablation studies under four modified settings. As shown in Table~\ref{tab:ablation}, removing the synthetic data and training only on the original dataset leads to the most significant performance drop across all benchmarks, demonstrating the high quality and strong utility of our synthesized data. Eliminating Dynamic Diagnosis and instead using the full set of skeletons from the original training set results in reduced performance, highlighting the effectiveness of model-specific augmentation. Disabling the Skeleton Generalizer and relying solely on error-prone skeletons limits structural diversity, resulting in performance decline. Finally, skipping Forward Verification and directly using unverified SQL-question pairs introduces semantic mismatches and hallucinations, negatively impacting performance. In conclusion, each component contributes meaningfully to the overall effectiveness of our method.

\begin{figure}[t]
\centering
  \includegraphics[width=\columnwidth]{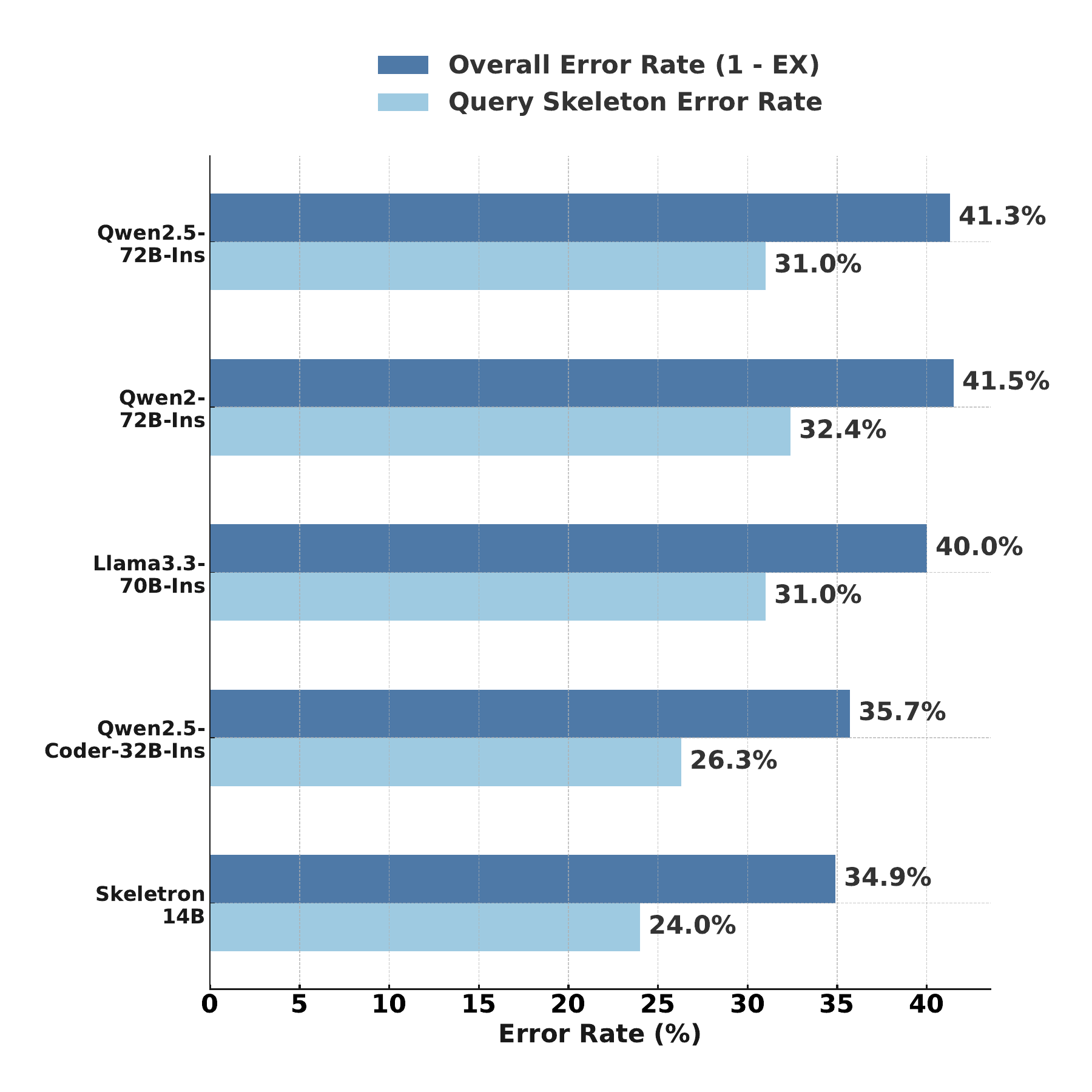}
  \caption{Comparison of overall error rate (1 - EX) and query skeleton error rate across different LLMs and Skeletron 14B on the BIRD Dev. The method for identifying query skeleton errors follows Section \ref{sec:error-detect}.}
  \label{fig:ske-err}
\end{figure}

\begin{figure}[t]
\centering
  \includegraphics[width=\columnwidth]{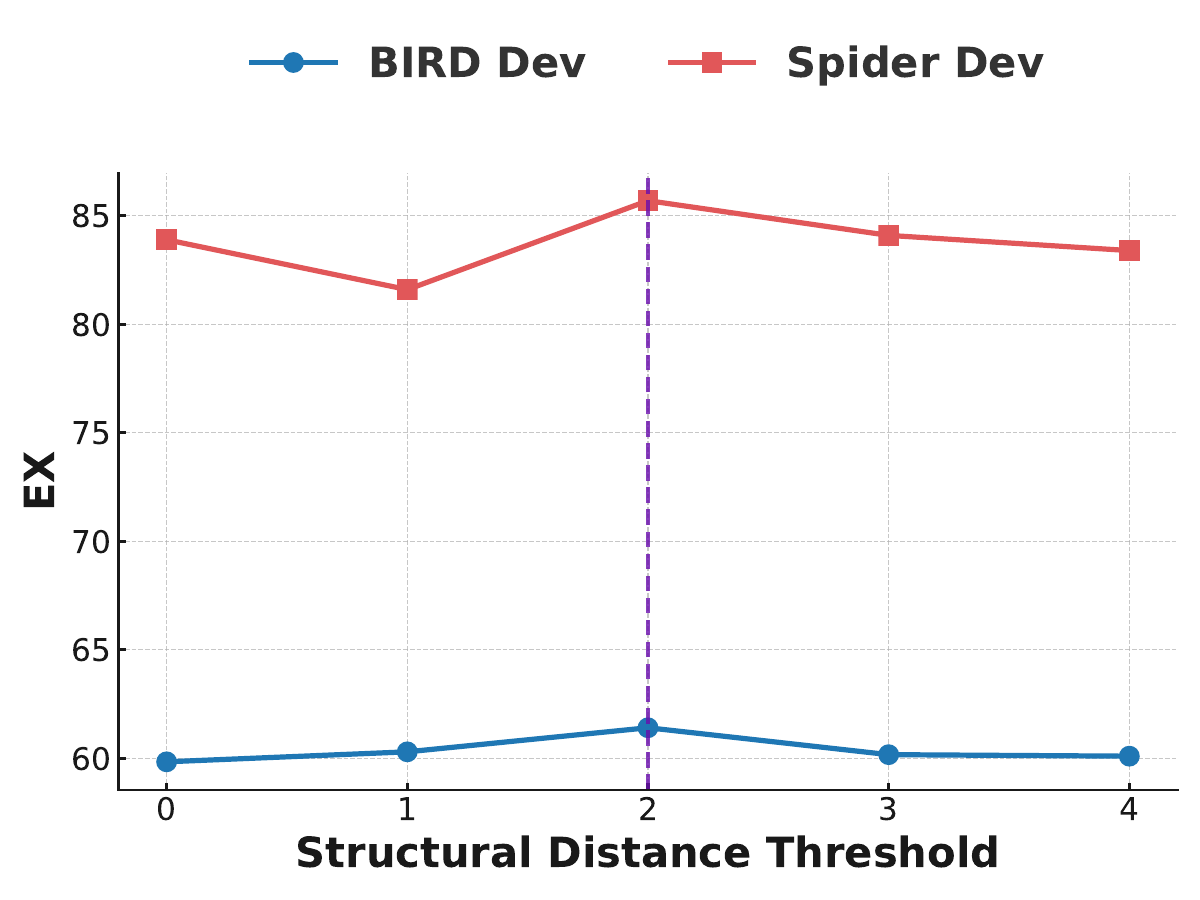}
  \caption{EX on the BIRD and Spider Dev sets under different structural edit distance thresholds used in the dynamic diagnosis step.}
  \label{fig:ts}
\end{figure}

\subsection{More Analysis}
\label{sec:analysis}
\paragraph{Can our method enhance LLMs' understanding of the skeletons of query languages?} We evaluate several state-of-the-art LLMs and our Skeletron 14B on the ability to predict correct query skeletons in Text-to-Query task. The results are shown in Figure~\ref{fig:ske-err}. We observe that even the most advanced open-source LLMs specialized in code still frequently fail to generate correct skeletons during inference. For instance, Qwen2.5-Coder-32B-Instruct achieves a 35.7\% overall error rate on the BIRD dev set, with 26.3\% of predictions exhibiting incorrect skeletons, accounting for 73.4\% of all errors. This indicates that current LLMs still fall short in reliably handling query languages. In contrast, Skeletron 14B not only reduces the overall error rate but also lowers the skeleton error rate to 24.0\%, demonstrating improved understanding of skeletons of query languages.

\paragraph{How to choose the structural distance threshold in dynamic diagnosis?}
We further investigate how the choice of threshold for structural similarity affects the effectiveness of dynamic diagnosis. As shown in Figure~\ref{fig:ts}, we evaluate model performance on the Text-to-SQL task using different threshold values. We find that setting the threshold too low can lead to overly strict error detection, mistakenly classifying semantically well-aligned predictions as skeleton errors and introducing noisy or unnecessary cases into the augmentation process. Conversely, a threshold that is too high (e.g., 4) may overlook genuinely misaligned skeletons, missing critical opportunities to strengthen the model’s weak points. Although this experiment is based on the Text-to-SQL task, it highlights a general principle for Text-to-Query: the criterion for detecting skeleton errors in dynamic diagnosis should strike a balance between strictness and leniency.

\section{Conclusion}
In this paper, we introduce and formally define the Text-to-Query task paradigm, unifying semantic parsing tasks across various query languages. We identified query skeletons as a critical and universal abstraction for analyzing model behaviors, diagnosing weaknesses, and guiding data synthesis. Based on this abstraction, we proposed a dynamic data augmentation framework that explicitly diagnoses model-specific structural weaknesses and generates targeted, high-quality training examples accordingly. Experimental results across four diverse Text-to-Query benchmarks demonstrated that our approach achieves state-of-the-art performance, even with a limited amount of synthesized training data. These findings not only highlight the efficiency and generality of our method but also lay a robust foundation for future unified research in the Text-to-Query Task.

\section*{Limitations}

Although our method demonstrates strong performance and generality across multiple query languages, there remain several limitations.

First, beyond the Text-to-SQL domain, the availability of high-quality datasets and standardized evaluation protocols remains limited. As a result, our experiments and baseline comparisons in other domains such as Text-to-Cypher and Text-to-nGQL are relatively constrained. We hope that future work will introduce more comprehensive datasets and unified evaluation settings to better assess our method.

Second, while our data augmentation framework is broadly applicable to different Text-to-Query tasks, the augmentation process is still performed independently for each task. The current setup does not support a unified model that can handle multiple query languages simultaneously. Developing a strong general Text-to-Query model remains an exciting direction for future work.

\section{Acknowledgments}
This work is supported by the Chinese NSF Major Research Plan (No.92270121) and the Fundamental Research Funds for the Central Universities 2232023D-19.

\bibliography{main}

@inproceedings{zhong-etal-2020-grounded,
    title = "Grounded Adaptation for Zero-shot Executable Semantic Parsing",
    author = "Zhong, Victor  and
      Lewis, Mike  and
      Wang, Sida I.  and
      Zettlemoyer, Luke",
    editor = "Webber, Bonnie  and
      Cohn, Trevor  and
      He, Yulan  and
      Liu, Yang",
    booktitle = "Proceedings of the 2020 Conference on Empirical Methods in Natural Language Processing (EMNLP)",
    month = nov,
    year = "2020",
    address = "Online",
    publisher = "Association for Computational Linguistics",
    url = "https://aclanthology.org/2020.emnlp-main.558/",
    doi = "10.18653/v1/2020.emnlp-main.558",
    pages = "6869--6882"
}

@inproceedings{zhong-etal-2025-synthet2c,
    title = "{S}ynthe{T}2{C}: Generating Synthetic Data for Fine-Tuning Large Language Models on the {T}ext2{C}ypher Task",
    author = "Zhong, Zijie  and
      Zhong, Linqing  and
      Sun, Zhaoze  and
      Jin, Qingyun  and
      Qin, Zengchang  and
      Zhang, Xiaofan",
    editor = "Rambow, Owen  and
      Wanner, Leo  and
      Apidianaki, Marianna  and
      Al-Khalifa, Hend  and
      Eugenio, Barbara Di  and
      Schockaert, Steven",
    booktitle = "Proceedings of the 31st International Conference on Computational Linguistics",
    month = jan,
    year = "2025",
    address = "Abu Dhabi, UAE",
    publisher = "Association for Computational Linguistics",
    url = "https://aclanthology.org/2025.coling-main.46/",
    pages = "672--692"
}

@inproceedings{tiwari-etal-2025-auto,
    title = "Auto-Cypher: Improving {LLM}s on Cypher generation via {LLM}-supervised generation-verification framework",
    author = "Tiwari, Aman  and
      Malay, Shiva Krishna Reddy  and
      Yadav, Vikas  and
      Hashemi, Masoud  and
      Madhusudhan, Sathwik Tejaswi",
    editor = "Chiruzzo, Luis  and
      Ritter, Alan  and
      Wang, Lu",
    booktitle = "Proceedings of the 2025 Conference of the Nations of the Americas Chapter of the Association for Computational Linguistics: Human Language Technologies (Volume 2: Short Papers)",
    month = apr,
    year = "2025",
    address = "Albuquerque, New Mexico",
    publisher = "Association for Computational Linguistics",
    url = "https://aclanthology.org/2025.naacl-short.53/",
    pages = "623--640",
    ISBN = "979-8-89176-190-2"
}

@inproceedings{cfg2,
    title = "Data Augmentation with Hierarchical {SQL}-to-Question Generation for Cross-domain Text-to-{SQL} Parsing",
    author = "Wu, Kun  and
      Wang, Lijie  and
      Li, Zhenghua  and
      Zhang, Ao  and
      Xiao, Xinyan  and
      Wu, Hua  and
      Zhang, Min  and
      Wang, Haifeng",
    editor = "Moens, Marie-Francine  and
      Huang, Xuanjing  and
      Specia, Lucia  and
      Yih, Scott Wen-tau",
    booktitle = "Proceedings of the 2021 Conference on Empirical Methods in Natural Language Processing",
    month = nov,
    year = "2021",
    address = "Online and Punta Cana, Dominican Republic",
    publisher = "Association for Computational Linguistics",
    url = "https://aclanthology.org/2021.emnlp-main.707/",
    doi = "10.18653/v1/2021.emnlp-main.707",
    pages = "8974--8983"
}

@inproceedings{cotq2,
    title = "Improving {LLM}-based {KGQA} for multi-hop Question Answering with implicit reasoning in few-shot examples",
    author = "Shah, Mili  and
      Cahoon, Joyce  and
      Milletari, Mirco  and
      Tian, Jing  and
      Psallidas, Fotis  and
      Mueller, Andreas  and
      Litombe, Nick",
    editor = "Biswas, Russa  and
      Kaffee, Lucie-Aim{\'e}e  and
      Agarwal, Oshin  and
      Minervini, Pasquale  and
      Singh, Sameer  and
      de Melo, Gerard",
    booktitle = "Proceedings of the 1st Workshop on Knowledge Graphs and Large Language Models (KaLLM 2024)",
    month = aug,
    year = "2024",
    address = "Bangkok, Thailand",
    publisher = "Association for Computational Linguistics",
    url = "https://aclanthology.org/2024.kallm-1.13/",
    doi = "10.18653/v1/2024.kallm-1.13",
    pages = "125--135"
}

@inproceedings{
cotq1,
title={{CHASE}-{SQL}: Multi-Path Reasoning and Preference Optimized Candidate Selection in Text-to-{SQL}},
author={Mohammadreza Pourreza and Hailong Li and Ruoxi Sun and Yeounoh Chung and Shayan Talaei and Gaurav Tarlok Kakkar and Yu Gan and Amin Saberi and Fatma Ozcan and Sercan O Arik},
booktitle={The Thirteenth International Conference on Learning Representations},
year={2025},
url={https://openreview.net/forum?id=CvGqMD5OtX}
}

@misc{decomp3,
      title={CHESS: Contextual Harnessing for Efficient SQL Synthesis}, 
      author={Shayan Talaei and Mohammadreza Pourreza and Yu-Chen Chang and Azalia Mirhoseini and Amin Saberi},
      year={2024},
      eprint={2405.16755},
      archivePrefix={arXiv},
      primaryClass={cs.LG},
      url={https://arxiv.org/abs/2405.16755}, 
}

@inproceedings{decomp2,
    title = "{MAC}-{SQL}: A Multi-Agent Collaborative Framework for Text-to-{SQL}",
    author = "Wang, Bing  and
      Ren, Changyu  and
      Yang, Jian  and
      Liang, Xinnian  and
      Bai, Jiaqi  and
      Chai, LinZheng  and
      Yan, Zhao  and
      Zhang, Qian-Wen  and
      Yin, Di  and
      Sun, Xing  and
      Li, Zhoujun",
    editor = "Rambow, Owen  and
      Wanner, Leo  and
      Apidianaki, Marianna  and
      Al-Khalifa, Hend  and
      Eugenio, Barbara Di  and
      Schockaert, Steven",
    booktitle = "Proceedings of the 31st International Conference on Computational Linguistics",
    month = jan,
    year = "2025",
    address = "Abu Dhabi, UAE",
    publisher = "Association for Computational Linguistics",
    url = "https://aclanthology.org/2025.coling-main.36/",
    pages = "540--557"
}

@misc{decomp1,
      title={C3: Zero-shot Text-to-SQL with ChatGPT}, 
      author={Xuemei Dong and Chao Zhang and Yuhang Ge and Yuren Mao and Yunjun Gao and lu Chen and Jinshu Lin and Dongfang Lou},
      year={2023},
      eprint={2307.07306},
      archivePrefix={arXiv},
      primaryClass={cs.CL},
      url={https://arxiv.org/abs/2307.07306}, 
}

@inproceedings{fewshot3,
    title = "Investigating Large Language Models for Text-to-{SPARQL} Generation",
    author = "D{'}Abramo, Jacopo  and
      Zugarini, Andrea  and
      Torroni, Paolo",
    editor = "Shi, Weijia  and
      Yu, Wenhao  and
      Asai, Akari  and
      Jiang, Meng  and
      Durrett, Greg  and
      Hajishirzi, Hannaneh  and
      Zettlemoyer, Luke",
    booktitle = "Proceedings of the 4th International Workshop on Knowledge-Augmented Methods for Natural Language Processing",
    month = may,
    year = "2025",
    address = "Albuquerque, New Mexico, USA",
    publisher = "Association for Computational Linguistics",
    url = "https://aclanthology.org/2025.knowledgenlp-1.5/",
    pages = "66--80",
    ISBN = "979-8-89176-229-9"
}

@inproceedings{fewshot2,
    title = "{ACT}-{SQL}: In-Context Learning for Text-to-{SQL} with Automatically-Generated Chain-of-Thought",
    author = "Zhang, Hanchong  and
      Cao, Ruisheng  and
      Chen, Lu  and
      Xu, Hongshen  and
      Yu, Kai",
    editor = "Bouamor, Houda  and
      Pino, Juan  and
      Bali, Kalika",
    booktitle = "Findings of the Association for Computational Linguistics: EMNLP 2023",
    month = dec,
    year = "2023",
    address = "Singapore",
    publisher = "Association for Computational Linguistics",
    url = "https://aclanthology.org/2023.findings-emnlp.227/",
    doi = "10.18653/v1/2023.findings-emnlp.227",
    pages = "3501--3532"
}

@inproceedings{fewshot1,
    title = "Enhancing Text-to-{SQL} Capabilities of Large Language Models: A Study on Prompt Design Strategies",
    author = "Nan, Linyong  and
      Zhao, Yilun  and
      Zou, Weijin  and
      Ri, Narutatsu  and
      Tae, Jaesung  and
      Zhang, Ellen  and
      Cohan, Arman  and
      Radev, Dragomir",
    editor = "Bouamor, Houda  and
      Pino, Juan  and
      Bali, Kalika",
    booktitle = "Findings of the Association for Computational Linguistics: EMNLP 2023",
    month = dec,
    year = "2023",
    address = "Singapore",
    publisher = "Association for Computational Linguistics",
    url = "https://aclanthology.org/2023.findings-emnlp.996/",
    doi = "10.18653/v1/2023.findings-emnlp.996",
    pages = "14935--14956"
}

@inproceedings{pourreza-rafiei-2024-dts,
    title = "{DTS}-{SQL}: Decomposed Text-to-{SQL} with Small Large Language Models",
    author = "Pourreza, Mohammadreza  and
      Rafiei, Davood",
    editor = "Al-Onaizan, Yaser  and
      Bansal, Mohit  and
      Chen, Yun-Nung",
    booktitle = "Findings of the Association for Computational Linguistics: EMNLP 2024",
    month = nov,
    year = "2024",
    address = "Miami, Florida, USA",
    publisher = "Association for Computational Linguistics",
    url = "https://aclanthology.org/2024.findings-emnlp.481/",
    doi = "10.18653/v1/2024.findings-emnlp.481",
    pages = "8212--8220"
}

@inproceedings{redsql,
author = {Li, Haoyang and Zhang, Jing and Li, Cuiping and Chen, Hong},
title = {RESDSQL: decoupling schema linking and skeleton parsing for text-to-SQL},
year = {2023},
isbn = {978-1-57735-880-0},
publisher = {AAAI Press},
url = {https://doi.org/10.1609/aaai.v37i11.26535},
doi = {10.1609/aaai.v37i11.26535},
booktitle = {Proceedings of the Thirty-Seventh AAAI Conference on Artificial Intelligence and Thirty-Fifth Conference on Innovative Applications of Artificial Intelligence and Thirteenth Symposium on Educational Advances in Artificial Intelligence},
articleno = {1466},
numpages = {9},
series = {AAAI'23/IAAI'23/EAAI'23}
}

@article{codes,
author = {Li, Haoyang and Zhang, Jing and Liu, Hanbing and Fan, Ju and Zhang, Xiaokang and Zhu, Jun and Wei, Renjie and Pan, Hongyan and Li, Cuiping and Chen, Hong},
title = {CodeS: Towards Building Open-source Language Models for Text-to-SQL},
year = {2024},
issue_date = {June 2024},
publisher = {Association for Computing Machinery},
address = {New York, NY, USA},
volume = {2},
number = {3},
url = {https://doi.org/10.1145/3654930},
doi = {10.1145/3654930},
journal = {Proc. ACM Manag. Data},
month = may,
articleno = {127},
numpages = {28}
}

@misc{grattafiori2024llama3herdmodels,
      title={The Llama 3 Herd of Models}, 
      author={Aaron Grattafiori and Abhimanyu Dubey and Abhinav Jauhri and Abhinav Pandey and Abhishek Kadian and Ahmad Al-Dahle and Aiesha Letman and Akhil Mathur and Alan Schelten and Alex Vaughan and Amy Yang and Angela Fan and Anirudh Goyal and Anthony Hartshorn and Aobo Yang and Archi Mitra and Archie Sravankumar and Artem Korenev and Arthur Hinsvark and Arun Rao and Aston Zhang and Aurelien Rodriguez and Austen Gregerson and Ava Spataru and Baptiste Roziere and Bethany Biron and Binh Tang and Bobbie Chern and Charlotte Caucheteux and Chaya Nayak and Chloe Bi and Chris Marra and Chris McConnell and Christian Keller and Christophe Touret and Chunyang Wu and Corinne Wong and Cristian Canton Ferrer and Cyrus Nikolaidis and Damien Allonsius and Daniel Song and Danielle Pintz and Danny Livshits and Danny Wyatt and David Esiobu and Dhruv Choudhary and Dhruv Mahajan and Diego Garcia-Olano and Diego Perino and Dieuwke Hupkes and Egor Lakomkin and Ehab AlBadawy and Elina Lobanova and Emily Dinan and Eric Michael Smith and Filip Radenovic and Francisco Guzmán and Frank Zhang and Gabriel Synnaeve and Gabrielle Lee and Georgia Lewis Anderson and Govind Thattai and Graeme Nail and Gregoire Mialon and Guan Pang and Guillem Cucurell and Hailey Nguyen and Hannah Korevaar and Hu Xu and Hugo Touvron and Iliyan Zarov and Imanol Arrieta Ibarra and Isabel Kloumann and Ishan Misra and Ivan Evtimov and Jack Zhang and Jade Copet and Jaewon Lee and Jan Geffert and Jana Vranes and Jason Park and Jay Mahadeokar and Jeet Shah and Jelmer van der Linde and Jennifer Billock and Jenny Hong and Jenya Lee and Jeremy Fu and Jianfeng Chi and Jianyu Huang and Jiawen Liu and Jie Wang and Jiecao Yu and Joanna Bitton and Joe Spisak and Jongsoo Park and Joseph Rocca and Joshua Johnstun and Joshua Saxe and Junteng Jia and Kalyan Vasuden Alwala and Karthik Prasad and Kartikeya Upasani and Kate Plawiak and Ke Li and Kenneth Heafield and Kevin Stone and Khalid El-Arini and Krithika Iyer and Kshitiz Malik and Kuenley Chiu and Kunal Bhalla and Kushal Lakhotia and Lauren Rantala-Yeary and Laurens van der Maaten and Lawrence Chen and Liang Tan and Liz Jenkins and Louis Martin and Lovish Madaan and Lubo Malo and Lukas Blecher and Lukas Landzaat and Luke de Oliveira and Madeline Muzzi and Mahesh Pasupuleti and Mannat Singh and Manohar Paluri and Marcin Kardas and Maria Tsimpoukelli and Mathew Oldham and Mathieu Rita and Maya Pavlova and Melanie Kambadur and Mike Lewis and Min Si and Mitesh Kumar Singh and Mona Hassan and Naman Goyal and Narjes Torabi and Nikolay Bashlykov and Nikolay Bogoychev and Niladri Chatterji and Ning Zhang and Olivier Duchenne and Onur Çelebi and Patrick Alrassy and Pengchuan Zhang and Pengwei Li and Petar Vasic and Peter Weng and Prajjwal Bhargava and Pratik Dubal and Praveen Krishnan and Punit Singh Koura and Puxin Xu and Qing He and Qingxiao Dong and Ragavan Srinivasan and Raj Ganapathy and Ramon Calderer and Ricardo Silveira Cabral and Robert Stojnic and Roberta Raileanu and Rohan Maheswari and Rohit Girdhar and Rohit Patel and Romain Sauvestre and Ronnie Polidoro and Roshan Sumbaly and Ross Taylor and Ruan Silva and Rui Hou and Rui Wang and Saghar Hosseini and Sahana Chennabasappa and Sanjay Singh and Sean Bell and Seohyun Sonia Kim and Sergey Edunov and Shaoliang Nie and Sharan Narang and Sharath Raparthy and Sheng Shen and Shengye Wan and Shruti Bhosale and Shun Zhang and Simon Vandenhende and Soumya Batra and Spencer Whitman and Sten Sootla and Stephane Collot and Suchin Gururangan and Sydney Borodinsky and Tamar Herman and Tara Fowler and Tarek Sheasha and Thomas Georgiou and Thomas Scialom and Tobias Speckbacher and Todor Mihaylov and Tong Xiao and Ujjwal Karn and Vedanuj Goswami and Vibhor Gupta and Vignesh Ramanathan and Viktor Kerkez and Vincent Gonguet and Virginie Do and Vish Vogeti and Vítor Albiero and Vladan Petrovic and Weiwei Chu and Wenhan Xiong and Wenyin Fu and Whitney Meers and Xavier Martinet and Xiaodong Wang and Xiaofang Wang and Xiaoqing Ellen Tan and Xide Xia and Xinfeng Xie and Xuchao Jia and Xuewei Wang and Yaelle Goldschlag and Yashesh Gaur and Yasmine Babaei and Yi Wen and Yiwen Song and Yuchen Zhang and Yue Li and Yuning Mao and Zacharie Delpierre Coudert and Zheng Yan and Zhengxing Chen and Zoe Papakipos and Aaditya Singh and Aayushi Srivastava and Abha Jain and Adam Kelsey and Adam Shajnfeld and Adithya Gangidi and Adolfo Victoria and Ahuva Goldstand and Ajay Menon and Ajay Sharma and Alex Boesenberg and Alexei Baevski and Allie Feinstein and Amanda Kallet and Amit Sangani and Amos Teo and Anam Yunus and Andrei Lupu and Andres Alvarado and Andrew Caples and Andrew Gu and Andrew Ho and Andrew Poulton and Andrew Ryan and Ankit Ramchandani and Annie Dong and Annie Franco and Anuj Goyal and Aparajita Saraf and Arkabandhu Chowdhury and Ashley Gabriel and Ashwin Bharambe and Assaf Eisenman and Azadeh Yazdan and Beau James and Ben Maurer and Benjamin Leonhardi and Bernie Huang and Beth Loyd and Beto De Paola and Bhargavi Paranjape and Bing Liu and Bo Wu and Boyu Ni and Braden Hancock and Bram Wasti and Brandon Spence and Brani Stojkovic and Brian Gamido and Britt Montalvo and Carl Parker and Carly Burton and Catalina Mejia and Ce Liu and Changhan Wang and Changkyu Kim and Chao Zhou and Chester Hu and Ching-Hsiang Chu and Chris Cai and Chris Tindal and Christoph Feichtenhofer and Cynthia Gao and Damon Civin and Dana Beaty and Daniel Kreymer and Daniel Li and David Adkins and David Xu and Davide Testuggine and Delia David and Devi Parikh and Diana Liskovich and Didem Foss and Dingkang Wang and Duc Le and Dustin Holland and Edward Dowling and Eissa Jamil and Elaine Montgomery and Eleonora Presani and Emily Hahn and Emily Wood and Eric-Tuan Le and Erik Brinkman and Esteban Arcaute and Evan Dunbar and Evan Smothers and Fei Sun and Felix Kreuk and Feng Tian and Filippos Kokkinos and Firat Ozgenel and Francesco Caggioni and Frank Kanayet and Frank Seide and Gabriela Medina Florez and Gabriella Schwarz and Gada Badeer and Georgia Swee and Gil Halpern and Grant Herman and Grigory Sizov and Guangyi and Zhang and Guna Lakshminarayanan and Hakan Inan and Hamid Shojanazeri and Han Zou and Hannah Wang and Hanwen Zha and Haroun Habeeb and Harrison Rudolph and Helen Suk and Henry Aspegren and Hunter Goldman and Hongyuan Zhan and Ibrahim Damlaj and Igor Molybog and Igor Tufanov and Ilias Leontiadis and Irina-Elena Veliche and Itai Gat and Jake Weissman and James Geboski and James Kohli and Janice Lam and Japhet Asher and Jean-Baptiste Gaya and Jeff Marcus and Jeff Tang and Jennifer Chan and Jenny Zhen and Jeremy Reizenstein and Jeremy Teboul and Jessica Zhong and Jian Jin and Jingyi Yang and Joe Cummings and Jon Carvill and Jon Shepard and Jonathan McPhie and Jonathan Torres and Josh Ginsburg and Junjie Wang and Kai Wu and Kam Hou U and Karan Saxena and Kartikay Khandelwal and Katayoun Zand and Kathy Matosich and Kaushik Veeraraghavan and Kelly Michelena and Keqian Li and Kiran Jagadeesh and Kun Huang and Kunal Chawla and Kyle Huang and Lailin Chen and Lakshya Garg and Lavender A and Leandro Silva and Lee Bell and Lei Zhang and Liangpeng Guo and Licheng Yu and Liron Moshkovich and Luca Wehrstedt and Madian Khabsa and Manav Avalani and Manish Bhatt and Martynas Mankus and Matan Hasson and Matthew Lennie and Matthias Reso and Maxim Groshev and Maxim Naumov and Maya Lathi and Meghan Keneally and Miao Liu and Michael L. Seltzer and Michal Valko and Michelle Restrepo and Mihir Patel and Mik Vyatskov and Mikayel Samvelyan and Mike Clark and Mike Macey and Mike Wang and Miquel Jubert Hermoso and Mo Metanat and Mohammad Rastegari and Munish Bansal and Nandhini Santhanam and Natascha Parks and Natasha White and Navyata Bawa and Nayan Singhal and Nick Egebo and Nicolas Usunier and Nikhil Mehta and Nikolay Pavlovich Laptev and Ning Dong and Norman Cheng and Oleg Chernoguz and Olivia Hart and Omkar Salpekar and Ozlem Kalinli and Parkin Kent and Parth Parekh and Paul Saab and Pavan Balaji and Pedro Rittner and Philip Bontrager and Pierre Roux and Piotr Dollar and Polina Zvyagina and Prashant Ratanchandani and Pritish Yuvraj and Qian Liang and Rachad Alao and Rachel Rodriguez and Rafi Ayub and Raghotham Murthy and Raghu Nayani and Rahul Mitra and Rangaprabhu Parthasarathy and Raymond Li and Rebekkah Hogan and Robin Battey and Rocky Wang and Russ Howes and Ruty Rinott and Sachin Mehta and Sachin Siby and Sai Jayesh Bondu and Samyak Datta and Sara Chugh and Sara Hunt and Sargun Dhillon and Sasha Sidorov and Satadru Pan and Saurabh Mahajan and Saurabh Verma and Seiji Yamamoto and Sharadh Ramaswamy and Shaun Lindsay and Shaun Lindsay and Sheng Feng and Shenghao Lin and Shengxin Cindy Zha and Shishir Patil and Shiva Shankar and Shuqiang Zhang and Shuqiang Zhang and Sinong Wang and Sneha Agarwal and Soji Sajuyigbe and Soumith Chintala and Stephanie Max and Stephen Chen and Steve Kehoe and Steve Satterfield and Sudarshan Govindaprasad and Sumit Gupta and Summer Deng and Sungmin Cho and Sunny Virk and Suraj Subramanian and Sy Choudhury and Sydney Goldman and Tal Remez and Tamar Glaser and Tamara Best and Thilo Koehler and Thomas Robinson and Tianhe Li and Tianjun Zhang and Tim Matthews and Timothy Chou and Tzook Shaked and Varun Vontimitta and Victoria Ajayi and Victoria Montanez and Vijai Mohan and Vinay Satish Kumar and Vishal Mangla and Vlad Ionescu and Vlad Poenaru and Vlad Tiberiu Mihailescu and Vladimir Ivanov and Wei Li and Wenchen Wang and Wenwen Jiang and Wes Bouaziz and Will Constable and Xiaocheng Tang and Xiaojian Wu and Xiaolan Wang and Xilun Wu and Xinbo Gao and Yaniv Kleinman and Yanjun Chen and Ye Hu and Ye Jia and Ye Qi and Yenda Li and Yilin Zhang and Ying Zhang and Yossi Adi and Youngjin Nam and Yu and Wang and Yu Zhao and Yuchen Hao and Yundi Qian and Yunlu Li and Yuzi He and Zach Rait and Zachary DeVito and Zef Rosnbrick and Zhaoduo Wen and Zhenyu Yang and Zhiwei Zhao and Zhiyu Ma},
      year={2024},
      eprint={2407.21783},
      archivePrefix={arXiv},
      primaryClass={cs.AI},
      url={https://arxiv.org/abs/2407.21783}, 
}

@misc{hui2024qwen25codertechnicalreport,
      title={Qwen2.5-Coder Technical Report}, 
      author={Binyuan Hui and Jian Yang and Zeyu Cui and Jiaxi Yang and Dayiheng Liu and Lei Zhang and Tianyu Liu and Jiajun Zhang and Bowen Yu and Keming Lu and Kai Dang and Yang Fan and Yichang Zhang and An Yang and Rui Men and Fei Huang and Bo Zheng and Yibo Miao and Shanghaoran Quan and Yunlong Feng and Xingzhang Ren and Xuancheng Ren and Jingren Zhou and Junyang Lin},
      year={2024},
      eprint={2409.12186},
      archivePrefix={arXiv},
      primaryClass={cs.CL},
      url={https://arxiv.org/abs/2409.12186}, 
}

@misc{yang2024qwen2technicalreport,
      title={Qwen2 Technical Report}, 
      author={An Yang and Baosong Yang and Binyuan Hui and Bo Zheng and Bowen Yu and Chang Zhou and Chengpeng Li and Chengyuan Li and Dayiheng Liu and Fei Huang and Guanting Dong and Haoran Wei and Huan Lin and Jialong Tang and Jialin Wang and Jian Yang and Jianhong Tu and Jianwei Zhang and Jianxin Ma and Jianxin Yang and Jin Xu and Jingren Zhou and Jinze Bai and Jinzheng He and Junyang Lin and Kai Dang and Keming Lu and Keqin Chen and Kexin Yang and Mei Li and Mingfeng Xue and Na Ni and Pei Zhang and Peng Wang and Ru Peng and Rui Men and Ruize Gao and Runji Lin and Shijie Wang and Shuai Bai and Sinan Tan and Tianhang Zhu and Tianhao Li and Tianyu Liu and Wenbin Ge and Xiaodong Deng and Xiaohuan Zhou and Xingzhang Ren and Xinyu Zhang and Xipin Wei and Xuancheng Ren and Xuejing Liu and Yang Fan and Yang Yao and Yichang Zhang and Yu Wan and Yunfei Chu and Yuqiong Liu and Zeyu Cui and Zhenru Zhang and Zhifang Guo and Zhihao Fan},
      year={2024},
      eprint={2407.10671},
      archivePrefix={arXiv},
      primaryClass={cs.CL},
      url={https://arxiv.org/abs/2407.10671}, 
}

@misc{halluc2,
      title={Hallucination is Inevitable: An Innate Limitation of Large Language Models}, 
      author={Ziwei Xu and Sanjay Jain and Mohan Kankanhalli},
      year={2025},
      eprint={2401.11817},
      archivePrefix={arXiv},
      primaryClass={cs.CL},
      url={https://arxiv.org/abs/2401.11817}, 
}

@article{halluc1,
   title={A Survey on Hallucination in Large Language Models: Principles, Taxonomy, Challenges, and Open Questions},
   volume={43},
   ISSN={1558-2868},
   url={http://dx.doi.org/10.1145/3703155},
   DOI={10.1145/3703155},
   number={2},
   journal={ACM Transactions on Information Systems},
   publisher={Association for Computing Machinery (ACM)},
   author={Huang, Lei and Yu, Weijiang and Ma, Weitao and Zhong, Weihong and Feng, Zhangyin and Wang, Haotian and Chen, Qianglong and Peng, Weihua and Feng, Xiaocheng and Qin, Bing and Liu, Ting},
   year={2025},
   month=jan, pages={1–55} }

@inproceedings{
xu2024wizardlm,
title={Wizard{LM}: Empowering Large Pre-Trained Language Models to Follow Complex Instructions},
author={Can Xu and Qingfeng Sun and Kai Zheng and Xiubo Geng and Pu Zhao and Jiazhan Feng and Chongyang Tao and Qingwei Lin and Daxin Jiang},
booktitle={The Twelfth International Conference on Learning Representations},
year={2024},
url={https://openreview.net/forum?id=CfXh93NDgH}
}

@misc{scalequest,
      title={Unleashing Reasoning Capability of LLMs via Scalable Question Synthesis from Scratch}, 
      author={Yuyang Ding and Xinyu Shi and Xiaobo Liang and Juntao Li and Qiaoming Zhu and Min Zhang},
      year={2024},
      eprint={2410.18693},
      archivePrefix={arXiv},
      primaryClass={cs.CL},
      url={https://arxiv.org/abs/2410.18693}, 
}

@article{changedistll,
author = {Fluri, Beat and Würsch, Michael and Pinzger, Martin and Gall, Harald},
year = {2007},
month = {12},
pages = {725-743},
title = {Change Distilling:Tree Differencing for Fine-Grained Source Code Change Extraction},
volume = {33},
journal = {Software Engineering, IEEE Transactions on},
doi = {10.1109/TSE.2007.70731}
}

@inproceedings{sqlast,
    title = "Improving Retrieval-augmented Text-to-{SQL} with {AST}-based Ranking and Schema Pruning",
    author = "Shen, Zhili  and
      Vougiouklis, Pavlos  and
      Diao, Chenxin  and
      Vyas, Kaustubh  and
      Ji, Yuanyi  and
      Pan, Jeff Z.",
    editor = "Al-Onaizan, Yaser  and
      Bansal, Mohit  and
      Chen, Yun-Nung",
    booktitle = "Proceedings of the 2024 Conference on Empirical Methods in Natural Language Processing",
    month = nov,
    year = "2024",
    address = "Miami, Florida, USA",
    publisher = "Association for Computational Linguistics",
    url = "https://aclanthology.org/2024.emnlp-main.449/",
    doi = "10.18653/v1/2024.emnlp-main.449",
    pages = "7865--7879"
}

@misc{liu2025surveynl2sqllargelanguage,
      title={A Survey of NL2SQL with Large Language Models: Where are we, and where are we going?}, 
      author={Xinyu Liu and Shuyu Shen and Boyan Li and Peixian Ma and Runzhi Jiang and Yuxin Zhang and Ju Fan and Guoliang Li and Nan Tang and Yuyu Luo},
      year={2025},
      eprint={2408.05109},
      archivePrefix={arXiv},
      primaryClass={cs.DB},
      url={https://arxiv.org/abs/2408.05109}, 
}

@INPROCEEDINGS{codeast2,
  author={Yang, Shouguo and Cheng, Long and Zeng, Yicheng and Lang, Zhe and Zhu, Hongsong and Shi, Zhiqiang},
  booktitle={2021 51st Annual IEEE/IFIP International Conference on Dependable Systems and Networks (DSN)}, 
  title={Asteria: Deep Learning-based AST-Encoding for Cross-platform Binary Code Similarity Detection}, 
  year={2021},
  volume={},
  number={},
  pages={224-236},
  doi={10.1109/DSN48987.2021.00036}}

@inproceedings{codeast1,
    title = "Revisiting Code Similarity Evaluation with Abstract Syntax Tree Edit Distance",
    author = "Song, Yewei  and
      Lothritz, Cedric  and
      Tang, Xunzhu  and
      Bissyand{\'e}, Tegawend{\'e}  and
      Klein, Jacques",
    editor = "Ku, Lun-Wei  and
      Martins, Andre  and
      Srikumar, Vivek",
    booktitle = "Proceedings of the 62nd Annual Meeting of the Association for Computational Linguistics (Volume 2: Short Papers)",
    month = aug,
    year = "2024",
    address = "Bangkok, Thailand",
    publisher = "Association for Computational Linguistics",
    url = "https://aclanthology.org/2024.acl-short.3/",
    doi = "10.18653/v1/2024.acl-short.3",
    pages = "38--46"
}

@inproceedings{ozsoy-etal-2025-text2cypher,
    title = "{T}ext2{C}ypher: Bridging Natural Language and Graph Databases",
    author = "Ozsoy, Makbule Gulcin  and
      Messallem, Leila  and
      Besga, Jon  and
      Minneci, Gianandrea",
    editor = "Gesese, Genet Asefa  and
      Sack, Harald  and
      Paulheim, Heiko  and
      Merono-Penuela, Albert  and
      Chen, Lihu",
    booktitle = "Proceedings of the Workshop on Generative AI and Knowledge Graphs (GenAIK)",
    month = jan,
    year = "2025",
    address = "Abu Dhabi, UAE",
    publisher = "International Committee on Computational Linguistics",
    url = "https://aclanthology.org/2025.genaik-1.11/",
    pages = "100--108"
}

@inproceedings{bird,
title={Can {LLM} Already Serve as A Database Interface? A {BI}g Bench for Large-Scale Database Grounded Text-to-{SQL}s},
author={Jinyang Li and Binyuan Hui and GE QU and Jiaxi Yang and Binhua Li and Bowen Li and Bailin Wang and Bowen Qin and Ruiying Geng and Nan Huo and Xuanhe Zhou and Chenhao Ma and Guoliang Li and Kevin Chang and Fei Huang and Reynold Cheng and Yongbin Li},
booktitle={Thirty-seventh Conference on Neural Information Processing Systems Datasets and Benchmarks Track},
year={2023},
url={https://openreview.net/forum?id=dI4wzAE6uV}
}

@inproceedings{yu-etal-2018-spider,
    title = "{S}pider: A Large-Scale Human-Labeled Dataset for Complex and Cross-Domain Semantic Parsing and Text-to-{SQL} Task",
    author = "Yu, Tao  and
      Zhang, Rui  and
      Yang, Kai  and
      Yasunaga, Michihiro  and
      Wang, Dongxu  and
      Li, Zifan  and
      Ma, James  and
      Li, Irene  and
      Yao, Qingning  and
      Roman, Shanelle  and
      Zhang, Zilin  and
      Radev, Dragomir",
    editor = "Riloff, Ellen  and
      Chiang, David  and
      Hockenmaier, Julia  and
      Tsujii, Jun{'}ichi",
    booktitle = "Proceedings of the 2018 Conference on Empirical Methods in Natural Language Processing",
    month = oct # "-" # nov,
    year = "2018",
    address = "Brussels, Belgium",
    publisher = "Association for Computational Linguistics",
    url = "https://aclanthology.org/D18-1425/",
    doi = "10.18653/v1/D18-1425",
    pages = "3911--3921"
}

@inproceedings{cot2,
title={Chain of Thought Prompting Elicits Reasoning in Large Language Models},
author={Jason Wei and Xuezhi Wang and Dale Schuurmans and Maarten Bosma and brian ichter and Fei Xia and Ed H. Chi and Quoc V Le and Denny Zhou},
booktitle={Advances in Neural Information Processing Systems},
editor={Alice H. Oh and Alekh Agarwal and Danielle Belgrave and Kyunghyun Cho},
year={2022},
url={https://openreview.net/forum?id=_VjQlMeSB_J}
}

@inproceedings{cot1,
author = {Kojima, Takeshi and Gu, Shixiang Shane and Reid, Machel and Matsuo, Yutaka and Iwasawa, Yusuke},
title = {Large language models are zero-shot reasoners},
year = {2022},
isbn = {9781713871088},
publisher = {Curran Associates Inc.},
address = {Red Hook, NY, USA},
booktitle = {Proceedings of the 36th International Conference on Neural Information Processing Systems},
articleno = {1613},
numpages = {15},
location = {New Orleans, LA, USA},
series = {NIPS '22}
}

@inproceedings{back-trans2,
    title = "Learning to Synthesize Data for Semantic Parsing",
    author = "Wang, Bailin  and
      Yin, Wenpeng  and
      Lin, Xi Victoria  and
      Xiong, Caiming",
    editor = "Toutanova, Kristina  and
      Rumshisky, Anna  and
      Zettlemoyer, Luke  and
      Hakkani-Tur, Dilek  and
      Beltagy, Iz  and
      Bethard, Steven  and
      Cotterell, Ryan  and
      Chakraborty, Tanmoy  and
      Zhou, Yichao",
    booktitle = "Proceedings of the 2021 Conference of the North American Chapter of the Association for Computational Linguistics: Human Language Technologies",
    month = jun,
    year = "2021",
    address = "Online",
    publisher = "Association for Computational Linguistics",
    url = "https://aclanthology.org/2021.naacl-main.220/",
    doi = "10.18653/v1/2021.naacl-main.220",
    pages = "2760--2766"
}

@inproceedings{back-trans1,
    title = "Importance of Synthesizing High-quality Data for Text-to-{SQL} Parsing",
    author = "Hu, Yiqun  and
      Zhao, Yiyun  and
      Jiang, Jiarong  and
      Lan, Wuwei  and
      Zhu, Henghui  and
      Chauhan, Anuj  and
      Li, Alexander Hanbo  and
      Pan, Lin  and
      Wang, Jun  and
      Hang, Chung-Wei  and
      Zhang, Sheng  and
      Guo, Jiang  and
      Dong, Mingwen  and
      Lilien, Joseph  and
      Ng, Patrick  and
      Wang, Zhiguo  and
      Castelli, Vittorio  and
      Xiang, Bing",
    editor = "Rogers, Anna  and
      Boyd-Graber, Jordan  and
      Okazaki, Naoaki",
    booktitle = "Findings of the Association for Computational Linguistics: ACL 2023",
    month = jul,
    year = "2023",
    address = "Toronto, Canada",
    publisher = "Association for Computational Linguistics",
    url = "https://aclanthology.org/2023.findings-acl.86/",
    doi = "10.18653/v1/2023.findings-acl.86",
    pages = "1327--1343"
}

@incollection{reynolds2007diagnostic,
  author    = {Reynolds, Cecil R. and Fletcher-Janzen, Elaine},
  title     = {Diagnostic Prescriptive Teaching},
  booktitle = {Encyclopedia of Special Education},
  editor    = {Reynolds, Cecil R. and Vannest, Kimberly J. and Fletcher-Janzen, Elaine},
  pages     = {772},
  year      = {2007},
  publisher = {John Wiley \& Sons}
}

@article{sciencebenchmark,
author = {Zhang, Yi and Deriu, Jan and Katsogiannis-Meimarakis, George and Kosten, Catherine and Koutrika, Georgia and Stockinger, Kurt},
title = {ScienceBenchmark: A Complex Real-World Benchmark for Evaluating Natural Language to SQL Systems},
year = {2023},
issue_date = {December 2023},
publisher = {VLDB Endowment},
volume = {17},
number = {4},
issn = {2150-8097},
url = {https://doi.org/10.14778/3636218.3636225},
doi = {10.14778/3636218.3636225},
journal = {Proc. VLDB Endow.},
month = dec,
pages = {685–698},
numpages = {14}
}

@misc{li2025omnisqlsynthesizinghighqualitytexttosql,
      title={OmniSQL: Synthesizing High-quality Text-to-SQL Data at Scale}, 
      author={Haoyang Li and Shang Wu and Xiaokang Zhang and Xinmei Huang and Jing Zhang and Fuxin Jiang and Shuai Wang and Tieying Zhang and Jianjun Chen and Rui Shi and Hong Chen and Cuiping Li},
      year={2025},
      eprint={2503.02240},
      archivePrefix={arXiv},
      primaryClass={cs.CL},
      url={https://arxiv.org/abs/2503.02240}, 
}

@inproceedings{sense,
    title = "Synthesizing Text-to-{SQL} Data from Weak and Strong {LLM}s",
    author = "Yang, Jiaxi  and
      Hui, Binyuan  and
      Yang, Min  and
      Yang, Jian  and
      Lin, Junyang  and
      Zhou, Chang",
    editor = "Ku, Lun-Wei  and
      Martins, Andre  and
      Srikumar, Vivek",
    booktitle = "Proceedings of the 62nd Annual Meeting of the Association for Computational Linguistics (Volume 1: Long Papers)",
    month = aug,
    year = "2024",
    address = "Bangkok, Thailand",
    publisher = "Association for Computational Linguistics",
    url = "https://aclanthology.org/2024.acl-long.425/",
    doi = "10.18653/v1/2024.acl-long.425",
    pages = "7864--7875"
}

@inproceedings{wang-etal-2024-improving-demonstration,
    title = "Improving Demonstration Diversity by Human-Free Fusing for Text-to-{SQL}",
    author = "Wang, Dingzirui  and
      Dou, Longxu  and
      Zhang, Xuanliang  and
      Zhu, Qingfu  and
      Che, Wanxiang",
    editor = "Al-Onaizan, Yaser  and
      Bansal, Mohit  and
      Chen, Yun-Nung",
    booktitle = "Findings of the Association for Computational Linguistics: EMNLP 2024",
    month = nov,
    year = "2024",
    address = "Miami, Florida, USA",
    publisher = "Association for Computational Linguistics",
    url = "https://aclanthology.org/2024.findings-emnlp.65/",
    doi = "10.18653/v1/2024.findings-emnlp.65",
    pages = "1193--1207"
}

@article{dail_sql,
    author  =   {Dawei Gao and
    Haibin Wang and
    Yaliang Li and
    Xiuyu Sun and
    Yichen Qian and
    Bolin Ding and
    Jingren Zhou},
    title   =   {Text-to-SQL Empowered by Large Language Models: A Benchmark Evaluation},
    journal =   {CoRR},
    volume  =   {abs/2308.15363},
    year    =   {2023}
}

@inproceedings{
pourreza2023dinsql,
title={{DIN}-{SQL}: Decomposed In-Context Learning of Text-to-{SQL} with Self-Correction},
author={Mohammadreza Pourreza and Davood Rafiei},
booktitle={Thirty-seventh Conference on Neural Information Processing Systems},
year={2023},
url={https://openreview.net/forum?id=p53QDxSIc5}
}

@misc{qwen2025qwen25technicalreport,
      title={Qwen2.5 Technical Report}, 
      author={Qwen and : and An Yang and Baosong Yang and Beichen Zhang and Binyuan Hui and Bo Zheng and Bowen Yu and Chengyuan Li and Dayiheng Liu and Fei Huang and Haoran Wei and Huan Lin and Jian Yang and Jianhong Tu and Jianwei Zhang and Jianxin Yang and Jiaxi Yang and Jingren Zhou and Junyang Lin and Kai Dang and Keming Lu and Keqin Bao and Kexin Yang and Le Yu and Mei Li and Mingfeng Xue and Pei Zhang and Qin Zhu and Rui Men and Runji Lin and Tianhao Li and Tianyi Tang and Tingyu Xia and Xingzhang Ren and Xuancheng Ren and Yang Fan and Yang Su and Yichang Zhang and Yu Wan and Yuqiong Liu and Zeyu Cui and Zhenru Zhang and Zihan Qiu},
      year={2025},
      eprint={2412.15115},
      archivePrefix={arXiv},
      primaryClass={cs.CL},
      url={https://arxiv.org/abs/2412.15115}, 
}

@misc{openai2024gpt4technicalreport,
      title={GPT-4 Technical Report}, 
      author={OpenAI and Josh Achiam and Steven Adler and Sandhini Agarwal and Lama Ahmad and Ilge Akkaya and Florencia Leoni Aleman and Diogo Almeida and Janko Altenschmidt and Sam Altman and Shyamal Anadkat and Red Avila and Igor Babuschkin and Suchir Balaji and Valerie Balcom and Paul Baltescu and Haiming Bao and Mohammad Bavarian and Jeff Belgum and Irwan Bello and Jake Berdine and Gabriel Bernadett-Shapiro and Christopher Berner and Lenny Bogdonoff and Oleg Boiko and Madelaine Boyd and Anna-Luisa Brakman and Greg Brockman and Tim Brooks and Miles Brundage and Kevin Button and Trevor Cai and Rosie Campbell and Andrew Cann and Brittany Carey and Chelsea Carlson and Rory Carmichael and Brooke Chan and Che Chang and Fotis Chantzis and Derek Chen and Sully Chen and Ruby Chen and Jason Chen and Mark Chen and Ben Chess and Chester Cho and Casey Chu and Hyung Won Chung and Dave Cummings and Jeremiah Currier and Yunxing Dai and Cory Decareaux and Thomas Degry and Noah Deutsch and Damien Deville and Arka Dhar and David Dohan and Steve Dowling and Sheila Dunning and Adrien Ecoffet and Atty Eleti and Tyna Eloundou and David Farhi and Liam Fedus and Niko Felix and Simón Posada Fishman and Juston Forte and Isabella Fulford and Leo Gao and Elie Georges and Christian Gibson and Vik Goel and Tarun Gogineni and Gabriel Goh and Rapha Gontijo-Lopes and Jonathan Gordon and Morgan Grafstein and Scott Gray and Ryan Greene and Joshua Gross and Shixiang Shane Gu and Yufei Guo and Chris Hallacy and Jesse Han and Jeff Harris and Yuchen He and Mike Heaton and Johannes Heidecke and Chris Hesse and Alan Hickey and Wade Hickey and Peter Hoeschele and Brandon Houghton and Kenny Hsu and Shengli Hu and Xin Hu and Joost Huizinga and Shantanu Jain and Shawn Jain and Joanne Jang and Angela Jiang and Roger Jiang and Haozhun Jin and Denny Jin and Shino Jomoto and Billie Jonn and Heewoo Jun and Tomer Kaftan and Łukasz Kaiser and Ali Kamali and Ingmar Kanitscheider and Nitish Shirish Keskar and Tabarak Khan and Logan Kilpatrick and Jong Wook Kim and Christina Kim and Yongjik Kim and Jan Hendrik Kirchner and Jamie Kiros and Matt Knight and Daniel Kokotajlo and Łukasz Kondraciuk and Andrew Kondrich and Aris Konstantinidis and Kyle Kosic and Gretchen Krueger and Vishal Kuo and Michael Lampe and Ikai Lan and Teddy Lee and Jan Leike and Jade Leung and Daniel Levy and Chak Ming Li and Rachel Lim and Molly Lin and Stephanie Lin and Mateusz Litwin and Theresa Lopez and Ryan Lowe and Patricia Lue and Anna Makanju and Kim Malfacini and Sam Manning and Todor Markov and Yaniv Markovski and Bianca Martin and Katie Mayer and Andrew Mayne and Bob McGrew and Scott Mayer McKinney and Christine McLeavey and Paul McMillan and Jake McNeil and David Medina and Aalok Mehta and Jacob Menick and Luke Metz and Andrey Mishchenko and Pamela Mishkin and Vinnie Monaco and Evan Morikawa and Daniel Mossing and Tong Mu and Mira Murati and Oleg Murk and David Mély and Ashvin Nair and Reiichiro Nakano and Rajeev Nayak and Arvind Neelakantan and Richard Ngo and Hyeonwoo Noh and Long Ouyang and Cullen O'Keefe and Jakub Pachocki and Alex Paino and Joe Palermo and Ashley Pantuliano and Giambattista Parascandolo and Joel Parish and Emy Parparita and Alex Passos and Mikhail Pavlov and Andrew Peng and Adam Perelman and Filipe de Avila Belbute Peres and Michael Petrov and Henrique Ponde de Oliveira Pinto and Michael and Pokorny and Michelle Pokrass and Vitchyr H. Pong and Tolly Powell and Alethea Power and Boris Power and Elizabeth Proehl and Raul Puri and Alec Radford and Jack Rae and Aditya Ramesh and Cameron Raymond and Francis Real and Kendra Rimbach and Carl Ross and Bob Rotsted and Henri Roussez and Nick Ryder and Mario Saltarelli and Ted Sanders and Shibani Santurkar and Girish Sastry and Heather Schmidt and David Schnurr and John Schulman and Daniel Selsam and Kyla Sheppard and Toki Sherbakov and Jessica Shieh and Sarah Shoker and Pranav Shyam and Szymon Sidor and Eric Sigler and Maddie Simens and Jordan Sitkin and Katarina Slama and Ian Sohl and Benjamin Sokolowsky and Yang Song and Natalie Staudacher and Felipe Petroski Such and Natalie Summers and Ilya Sutskever and Jie Tang and Nikolas Tezak and Madeleine B. Thompson and Phil Tillet and Amin Tootoonchian and Elizabeth Tseng and Preston Tuggle and Nick Turley and Jerry Tworek and Juan Felipe Cerón Uribe and Andrea Vallone and Arun Vijayvergiya and Chelsea Voss and Carroll Wainwright and Justin Jay Wang and Alvin Wang and Ben Wang and Jonathan Ward and Jason Wei and CJ Weinmann and Akila Welihinda and Peter Welinder and Jiayi Weng and Lilian Weng and Matt Wiethoff and Dave Willner and Clemens Winter and Samuel Wolrich and Hannah Wong and Lauren Workman and Sherwin Wu and Jeff Wu and Michael Wu and Kai Xiao and Tao Xu and Sarah Yoo and Kevin Yu and Qiming Yuan and Wojciech Zaremba and Rowan Zellers and Chong Zhang and Marvin Zhang and Shengjia Zhao and Tianhao Zheng and Juntang Zhuang and William Zhuk and Barret Zoph},
      year={2024},
      eprint={2303.08774},
      archivePrefix={arXiv},
      primaryClass={cs.CL},
      url={https://arxiv.org/abs/2303.08774}, 
}

@inproceedings{popescu-etal-2004-modern,
    title = "Modern Natural Language Interfaces to Databases: Composing Statistical Parsing with Semantic Tractability",
    author = "Popescu, Ana-Maria  and
      Armanasu, Alex  and
      Etzioni, Oren  and
      Ko, David  and
      Yates, Alexander",
    booktitle = "{COLING} 2004: Proceedings of the 20th International Conference on Computational Linguistics",
    month = "aug 23–aug 27",
    year = "2004",
    address = "Geneva, Switzerland",
    publisher = "COLING",
    url = "https://aclanthology.org/C04-1021/",
    pages = "141--147"
}

@article{wikidata,
  author       = {Victor Zhong and
                  Caiming Xiong and
                  Richard Socher},
  title        = {Seq2SQL: Generating Structured Queries from Natural Language using
                  Reinforcement Learning},
  journal      = {CoRR},
  volume       = {abs/1709.00103},
  year         = {2017},
  url          = {http://arxiv.org/abs/1709.00103},
  eprinttype    = {arXiv},
  eprint       = {1709.00103},
  bibsource    = {dblp computer science bibliography, https://dblp.org}
}

@inproceedings{spccql,
author = {Guo, Aibo and Li, Xinyi and Xiao, Guanchen and Tan, Zhen and Zhao, Xiang},
title = {SpCQL: A Semantic Parsing Dataset for Converting Natural Language into Cypher},
year = {2022},
isbn = {9781450392365},
publisher = {Association for Computing Machinery},
address = {New York, NY, USA},
url = {https://doi.org/10.1145/3511808.3557703},
doi = {10.1145/3511808.3557703},
booktitle = {Proceedings of the 31st ACM International Conference on Information \& Knowledge Management},
pages = {3973–3977},
numpages = {5},
location = {Atlanta, GA, USA},
series = {CIKM '22}
}

@inproceedings{nl2gql,
    title = "$R^3$-{NL}2{GQL}: A Model Coordination and Knowledge Graph Alignment Approach for {NL}2{GQL}",
    author = "Zhou, Yuhang  and
      He, Yu  and
      Tian, Siyu  and
      Ni, Yuchen  and
      Yin, Zhangyue  and
      Liu, Xiang  and
      Ji, Chuanjun  and
      Liu, Sen  and
      Qiu, Xipeng  and
      Ye, Guangnan  and
      Chai, Hongfeng",
    editor = "Al-Onaizan, Yaser  and
      Bansal, Mohit  and
      Chen, Yun-Nung",
    booktitle = "Findings of the Association for Computational Linguistics: EMNLP 2024",
    month = nov,
    year = "2024",
    address = "Miami, Florida, USA",
    publisher = "Association for Computational Linguistics",
    url = "https://aclanthology.org/2024.findings-emnlp.800/",
    doi = "10.18653/v1/2024.findings-emnlp.800",
    pages = "13679--13692"
}

\appendix
\section{Schema Components}
\subsection{SQLite Database}
The schema of an SQLite database is organized in the form of DDL statements, including table names, column names, column types, optional column comments, optional sample values, and primary/foreign key constraints. Column comments consist of column descriptions and value illustrations, which are only available in the BIRD dataset. Therefore, they are included exclusively in the BIRD schemas. An example schema is shown in Figure~\ref{fig:sqlite_schema}.

During the SFT stage, we provide as much contextual information as possible for training. Thus, the schema includes column comments and 3 sample values for each column. However, due to GPU memory limitations, we restrict the input length to below 8192 characters, if the input exceeds this limit, we prioritize retaining the schemas of the gold tables corresponding to the SQL, while discarding the remaining table schemas.

In the data synthesis stage, for the same purpose of providing LLMs with sufficient information to better accomplish the task, 
we use the complete schema with all available content, except that excessively long sample values in some columns are selectively omitted.

In the inference stage, following the common evaluation setup \citep{bird,yu-etal-2018-spider,pourreza2023dinsql}, 
we exclude column comments from the schema and retain only 3 sample values per column along with other mandatory elements.
\begin{figure}[t]
\centering
\includegraphics[width=1\linewidth] {figures/appendix/sqlite\_schema.pdf}
\caption{An example of SQLite database schema.}
\label{fig:sqlite_schema}
\end{figure}

\subsection{Neo4j Database}
The Text2Cypher dataset \citep{ozsoy-etal-2025-text2cypher} already includes the corresponding database schemas, which we directly use. Each database schema consists of nodes, node properties (with types and sample values), and relation types. An example of a Neo4j database schema is shown in Figure~\ref{fig:neo4j_schema}.
\begin{figure}[t]
\centering
\includegraphics[width=1\linewidth] {figures/appendix/neo4j\_schema.pdf}
\caption{An example of Neo4j database schema.}
\label{fig:neo4j_schema}
\end{figure}

\subsection{NebulaGraph Database}
For the schema of NebulaGraph databases, we follow the setup of \citet{nl2gql} and organize the graph schema using Python code, which ensures semantic integrity across entities, relationships, and attributes while minimizing information loss.

Specifically, the code-structured schema encodes the graph in terms of Tags and Edges, where Python constructs are employed to provide detailed and precise descriptions: (1) concepts are defined as Python classes; (2) class annotations offer explanatory details; (3) class inheritance captures hierarchical relations; and (4) initialization functions specify the attributes of tags or edges. Figure~\ref{fig:nebula_schema} illustrates an example of such a graph schema.
\begin{figure}[t]
\centering
\includegraphics[width=1\linewidth] {figures/appendix/nebula\_schema.pdf}
\caption{An example of NebulaGraph database schema.}
\label{fig:nebula_schema}
\end{figure}

\section{Implementation of Structural Similarity Measure}
\label{app:measure}
For the computation of AST-based structural distance, we leverage SQLGlot\footnote{\url{https://github.com/tobymao/sqlglot}}, a comprehensive and generic SQL parser. SQLGlot provides an implementation of the Change Distiller algorithm, which computes the minimal set of edit operations required to transform one SQL AST into another. Further details of this implementation can be found in its documentation\footnote{\url{https://github.com/tobymao/sqlglot/blob/main/posts/sql_diff.md}}. For token-based structural distance, we simply split queries into tokens using whitespace as the delimiter.

\section{Details on Skeleton Generalizer}
\label{app:generalizer}
We fine-tune Qwen2.5-Coder-14B-Instruct with the error-prone skeletons obtained from the dynamic diagnosis step to derive a Skeleton Generalizer. Inspired by prior work \citep{xu2024wizardlm,scalequest}, we provide only a partial prefix of the LLM’s instruction template to guide the model in generating the corresponding skeletons. Instruction-tuned LLMs such as Qwen2.5-Coder-14B-Instruct have already learned to produce responses based on question-answer pairs (e.g., \textit{"<|im\_start|>User: \{instruction\}<|im\_end|>\textbackslash n<|im\_start|>Assistant: \{output\}<im\_end>"}). Since in our setting the model is only used for skeleton synthesis without any specific user questions, and query statements are more likely than questions to appear in the answer position during instruction tuning, we adopt the answer part of the instruction template (i.e., \textit{"<|im\_start|>Assistant:"}) to guide skeleton generation. This setup encourages more diverse skeletons. Nevertheless, to further promote the generation of error-prone skeletons and suppress unrelated content, additional fine-tuning is required.  

\begin{figure}[t]
\centering
\includegraphics[width=1\linewidth] {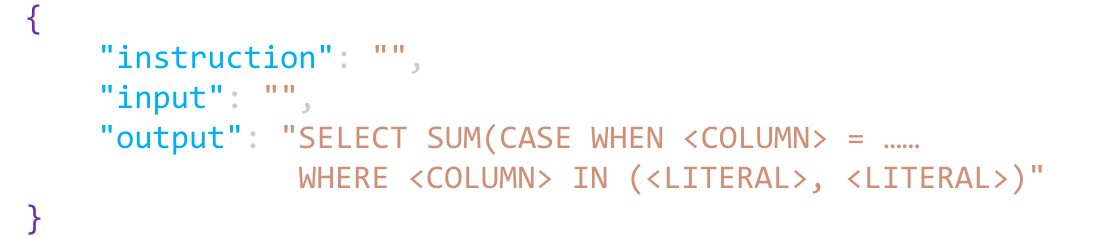}
\caption{An example from the fine-tuning dataset in Alpaca format.}
\label{fig:generalizer}
\end{figure}

Concretely, during fine-tuning we adjust the instruction template to \textit{"<|im\_start|>Assistant: \{output\}<im\_end>"}. An example from the fine-tuning dataset in Alpaca format is shown in Figure~\ref{fig:generalizer}. At inference time, we use the same instruction template, and the model can directly output diverse skeletons without requiring any explicit instruction.

\section{Skeletons Extraction}
To extract the skeletons of SQL queries, we employ SQLGlot to parse SQL queries into ASTs. We then traverse the ASTs to identify all tables, columns, and literals, replacing them with corresponding placeholders to obtain the SQL skeletons. For other query languages such as Cypher and nGQL, due to the lack of powerful open-source parsers, we predefine skeleton extraction rules and leverage LLMs to accomplish the extraction. The extraction rules for Cypher skeletons are illustrated in Figure~\ref{fig:e_cypher}, and those for nGQL skeletons are shown in Figure~\ref{fig:e_ngql}.
\begin{figure}[t]
\centering
\includegraphics[width=1\linewidth] {figures/appendix/extraction\_cypher.pdf}
\caption{Extraction rules for Cypher skeletons.}
\label{fig:e_cypher}
\end{figure}

\begin{figure}[t]
\centering
\includegraphics[width=1\linewidth] {figures/appendix/extraction\_ngql.pdf}
\caption{Extraction rules for nGQL skeletons.}
\label{fig:e_ngql}
\end{figure}

\section{Impact of Teacher Models}
To examine how our method performs with teacher models of different capacities, we conducted experiments using two smaller models, Qwen2.5-14B-Instruct and Qwen2.5-32B-Instruct, as alternatives to the original teacher model Qwen2.5-72B-Instruct used in the paper. The base model we used is Qwen2.5-Coder-7B and we evaluate it on the BIRD dataset. As shown in Table~\ref{teacher}, while using a weaker teacher does lead to a slight drop in performance, the overall decline is modest, indicating the robustness of our method to the choice of teacher model. At the same time, stronger teacher models do yield better results, suggesting that, when resources permit, such as using larger open-source models or even proprietary ones, the benefits of our method can be further amplified.

\begin{table}[t]
\scriptsize
\centering
\begin{tabular}{l|*{4}{>{\centering\arraybackslash}p{0.8cm}}}
\toprule
\multicolumn{1}{c|}{\textbf{Teacher Model}} & \textbf{simple} & \textbf{moderate} & \textbf{challenging} & \textbf{total} \\ \midrule
Qwen2.5-14B-Instruct                        & 64.8            & 50.2              & 42.1                 & 58.2           \\
Qwen2.5-32B-Instruct                        & 65.2            & 50.4              & 43.5                 & 58.7           \\
Qwen2.5-72B-Instruct                        & 67.6            & 53.5              & 47.6                 & 61.4           \\ \bottomrule
\end{tabular}
\caption{EX performance variations across different teacher models, evaluated on the BIRD dev set using Qwen2.5-Coder-7B as the target model.}
\label{teacher}
\end{table}

\newpage
\null
\newpage
\begin{figure*}[t]
\centering
\includegraphics[width=1\linewidth] {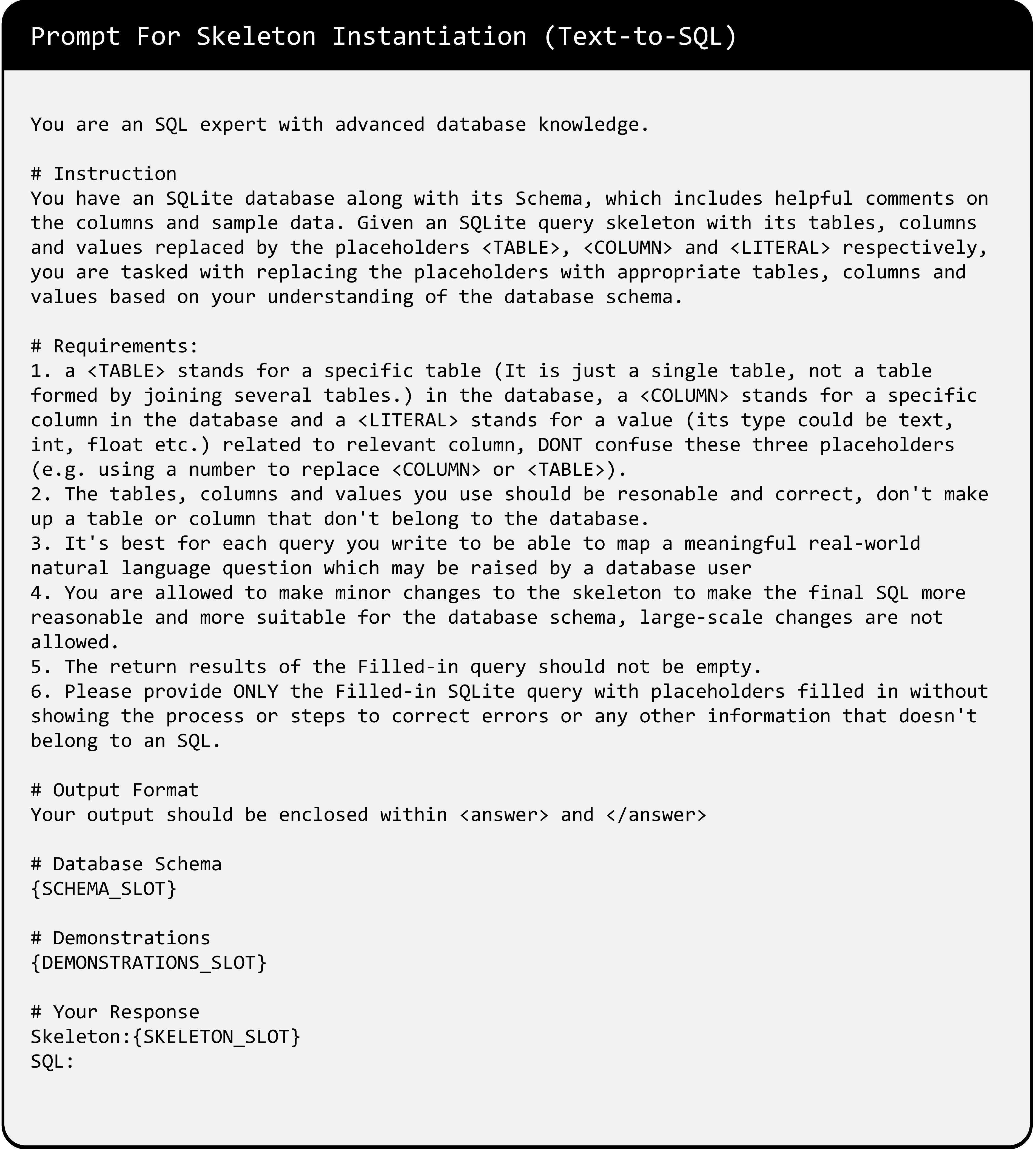}
\caption{Prompt for Skeleton Instantiation in the Data Synthesis Pipeline of Text-to-SQL.}
\end{figure*}

\newpage
\begin{figure*}[t]
\centering
\includegraphics[width=1\linewidth] {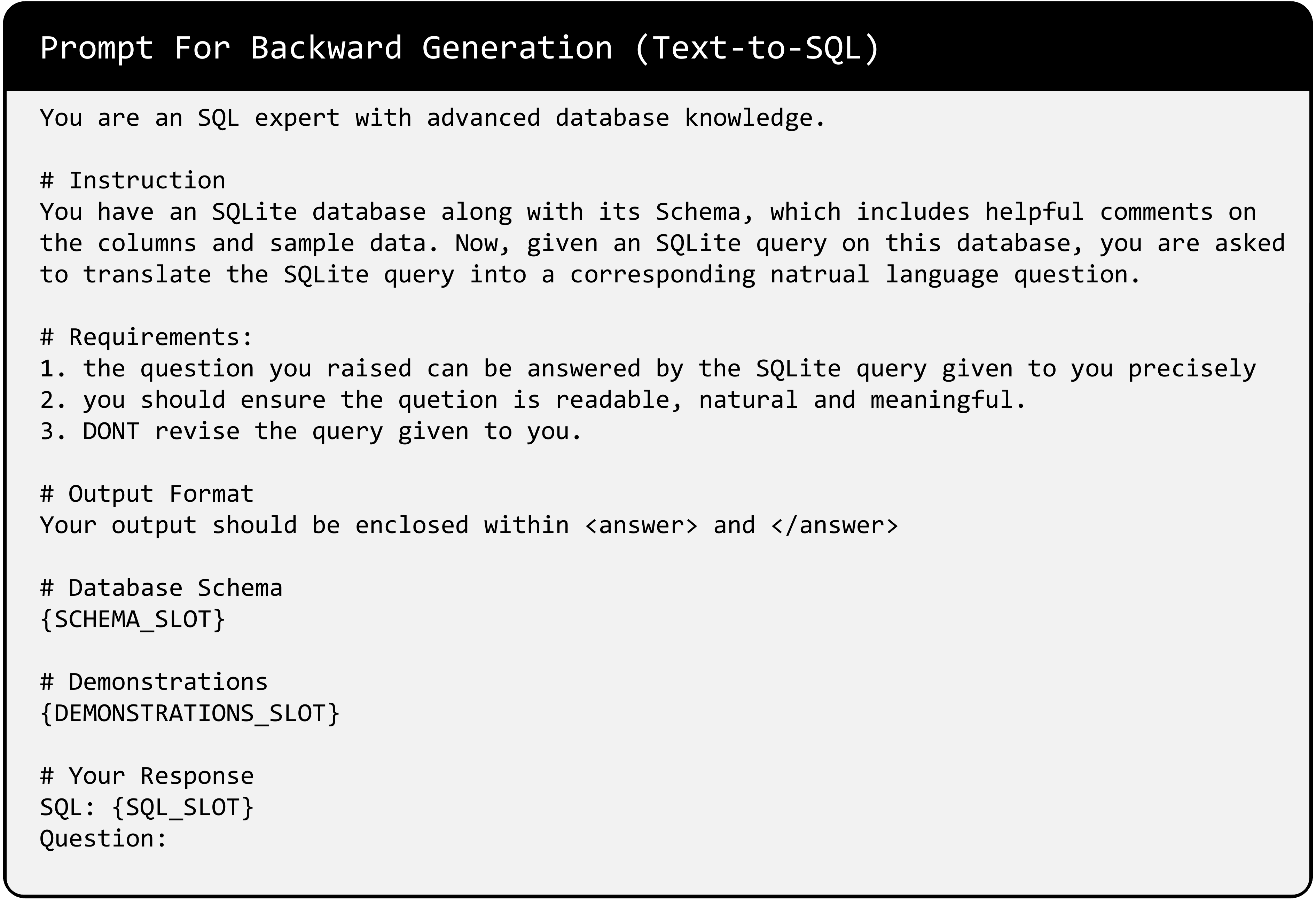}
\caption{Prompt for Backward Generation in the Data Synthesis Pipeline of Text-to-SQL.}
\end{figure*}

\newpage
\begin{figure*}[t]
\centering
\includegraphics[width=1\linewidth] {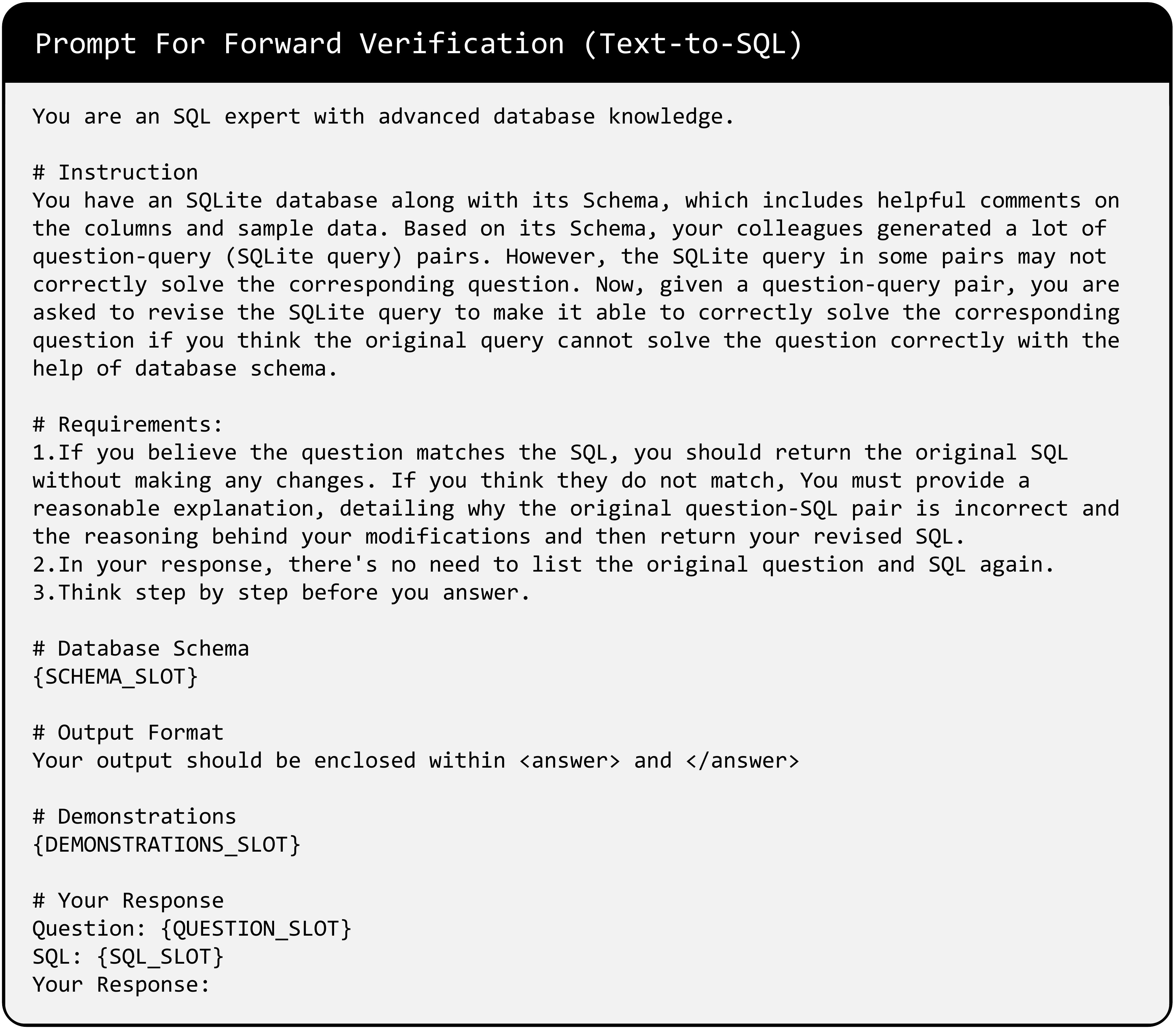}
\caption{Prompt for Forward Verification in the Data Synthesis Pipeline of Text-to-SQL.}
\end{figure*}

\newpage
\begin{figure*}[t]
\centering
\includegraphics[width=1\linewidth] {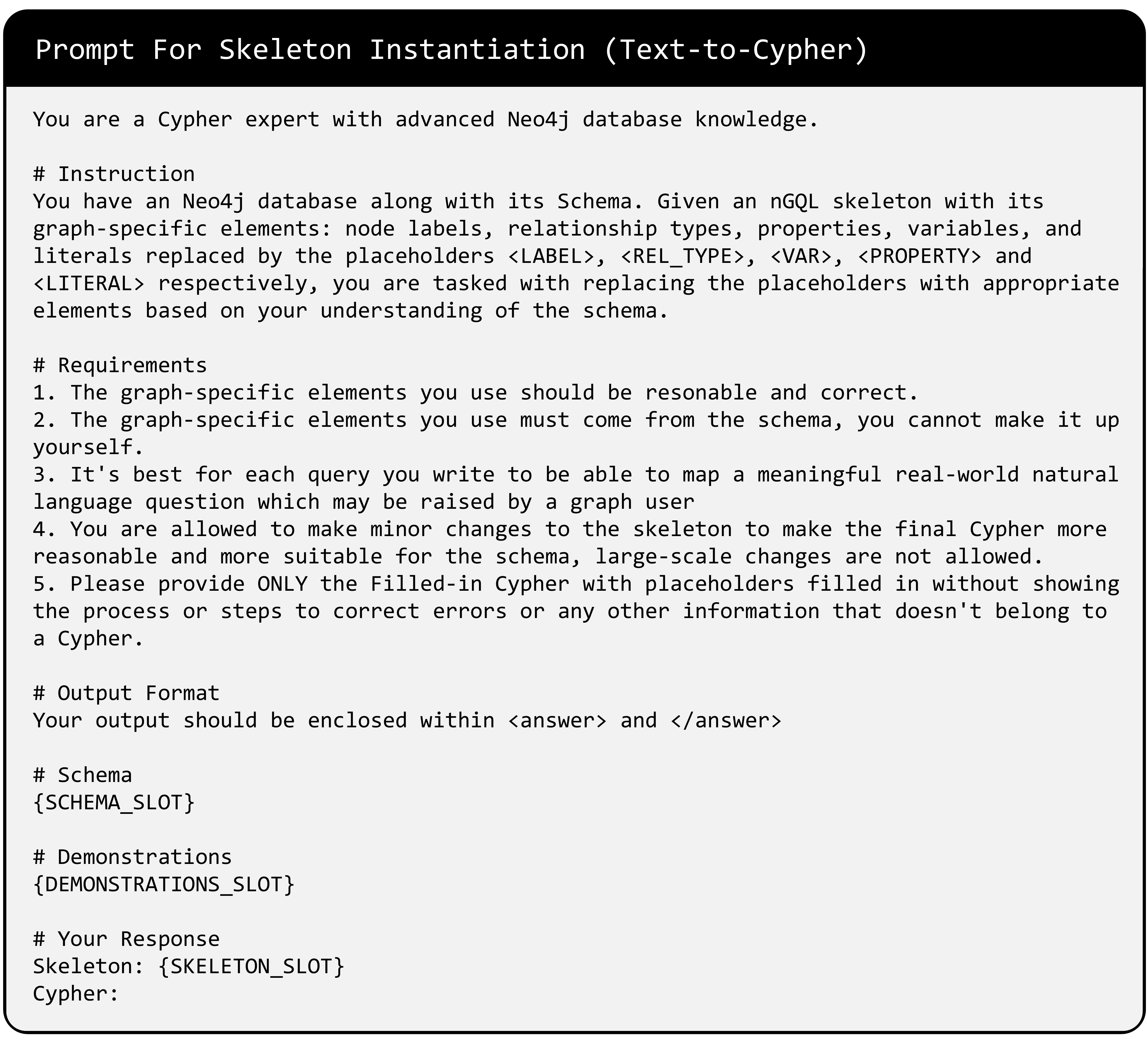}
\caption{Prompt for Skeleton Instantiation in the Data Synthesis Pipeline of Text-to-Cypher.}
\end{figure*}

\newpage
\begin{figure*}[t]
\centering
\includegraphics[width=1\linewidth] {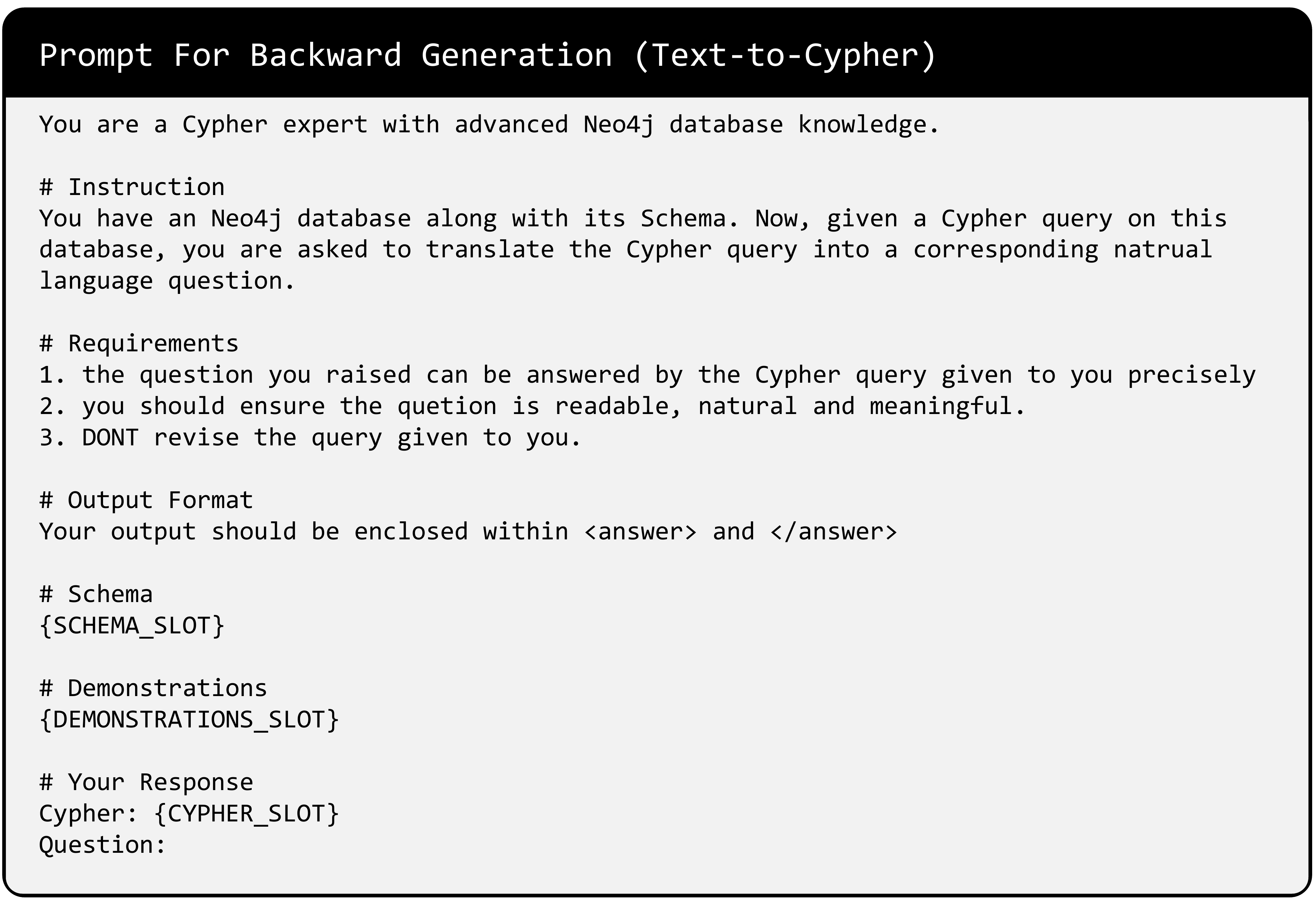}
\caption{Prompt for Backward Generation in the Data Synthesis Pipeline of Text-to-Cypher.}
\end{figure*}

\newpage
\begin{figure*}[t]
\centering
\includegraphics[width=1\linewidth] {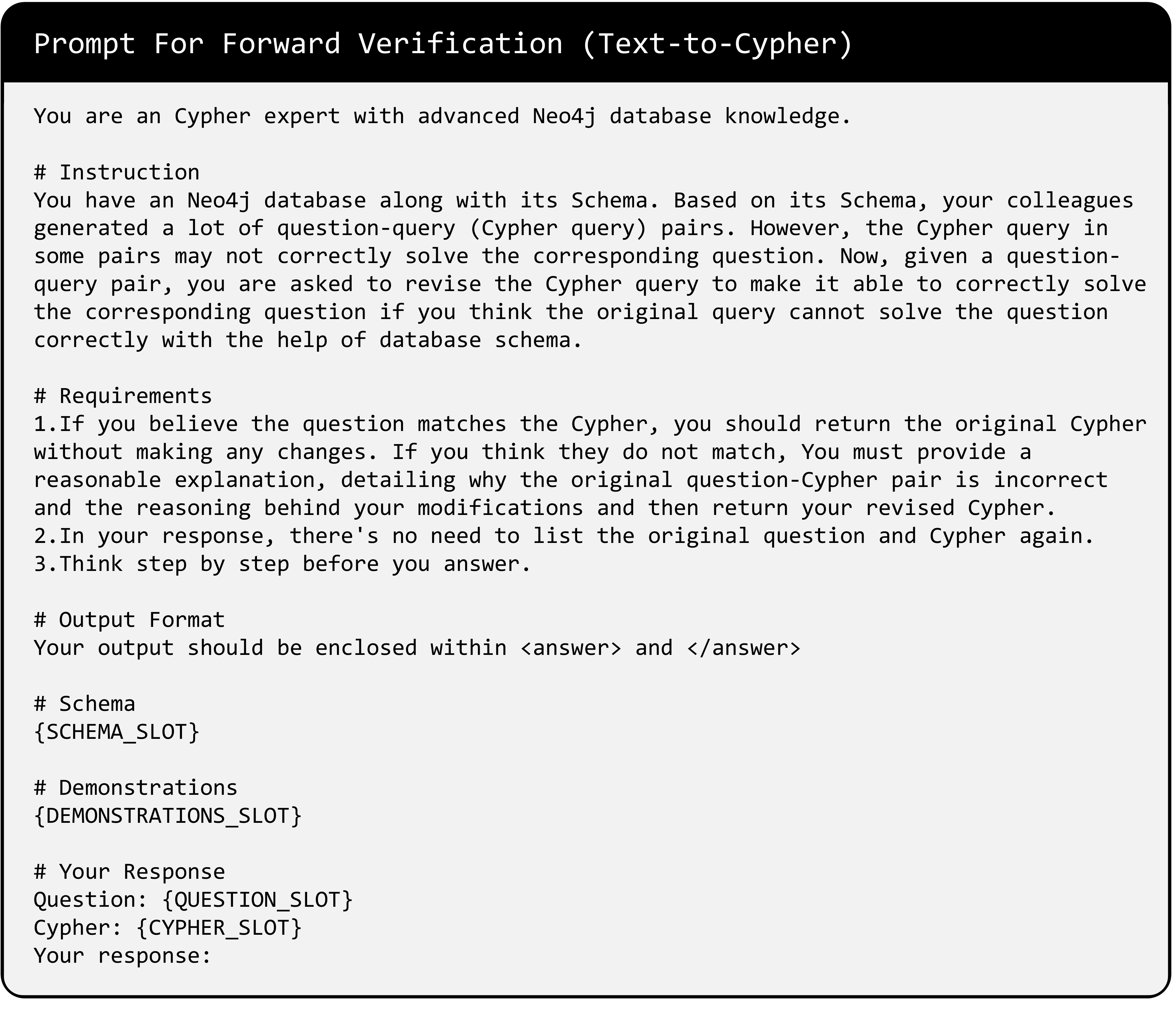}
\caption{Prompt for Forward Verification in the Data Synthesis Pipeline of Text-to-Cypher.}
\end{figure*}

\newpage
\begin{figure*}[t]
\centering
\includegraphics[width=1\linewidth] {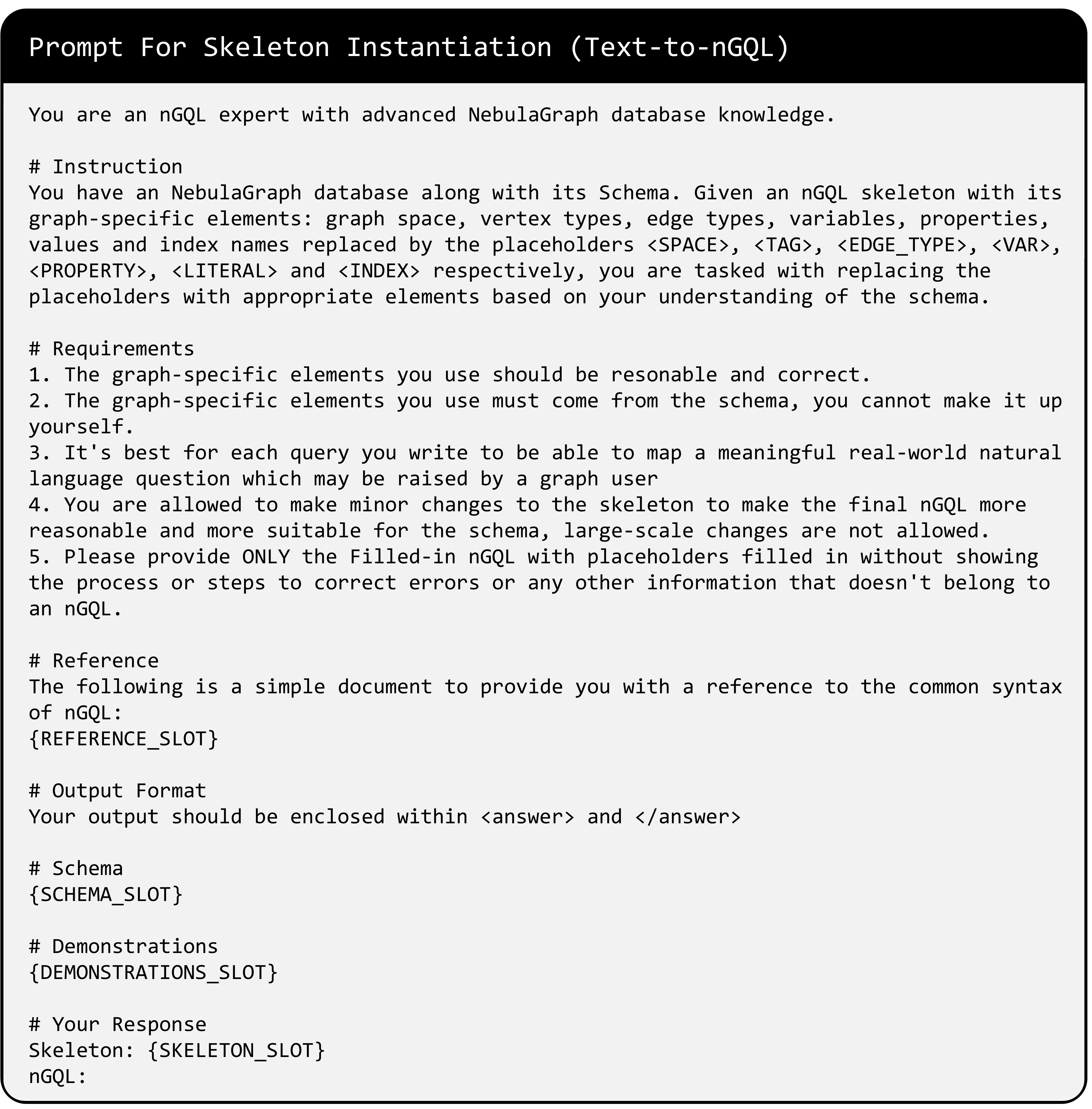}
\caption{Prompt for Skeleton Instantiation in the Data Synthesis Pipeline of Text-to-nGQL. The \textit{\{REFERENCE\_SLOT\}} part will be replaced with the Code-Structured Skeleton for GQL (including the framework of nGQL keywords) described in \citet{nl2gql}.}
\end{figure*}

\newpage
\begin{figure*}[t]
\centering
\includegraphics[width=1\linewidth] {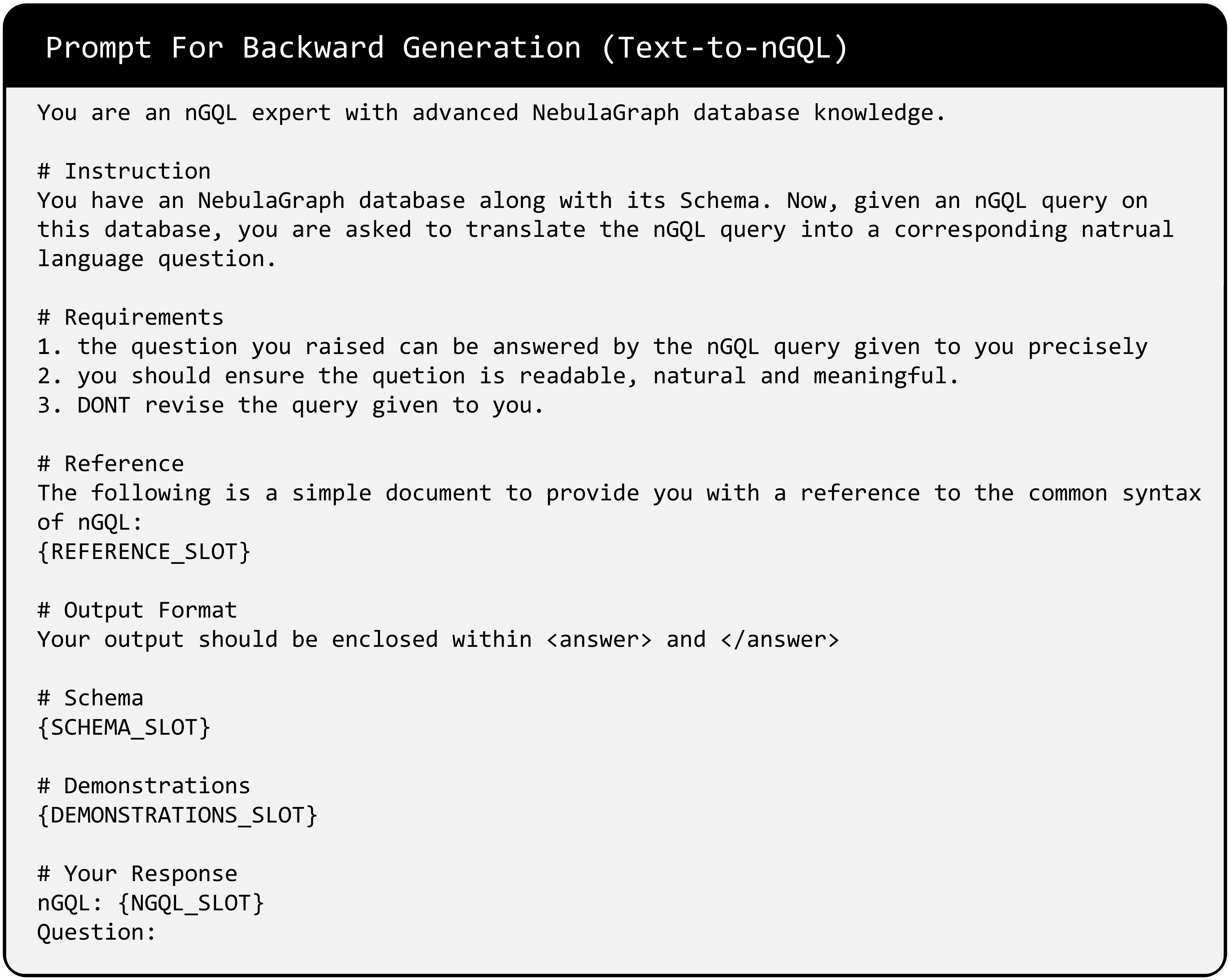}
\caption{Prompt for Backward Generation in the Data Synthesis Pipeline of Text-to-nGQL.}
\end{figure*}

\newpage
\begin{figure*}[t]
\centering
\includegraphics[width=1\linewidth] {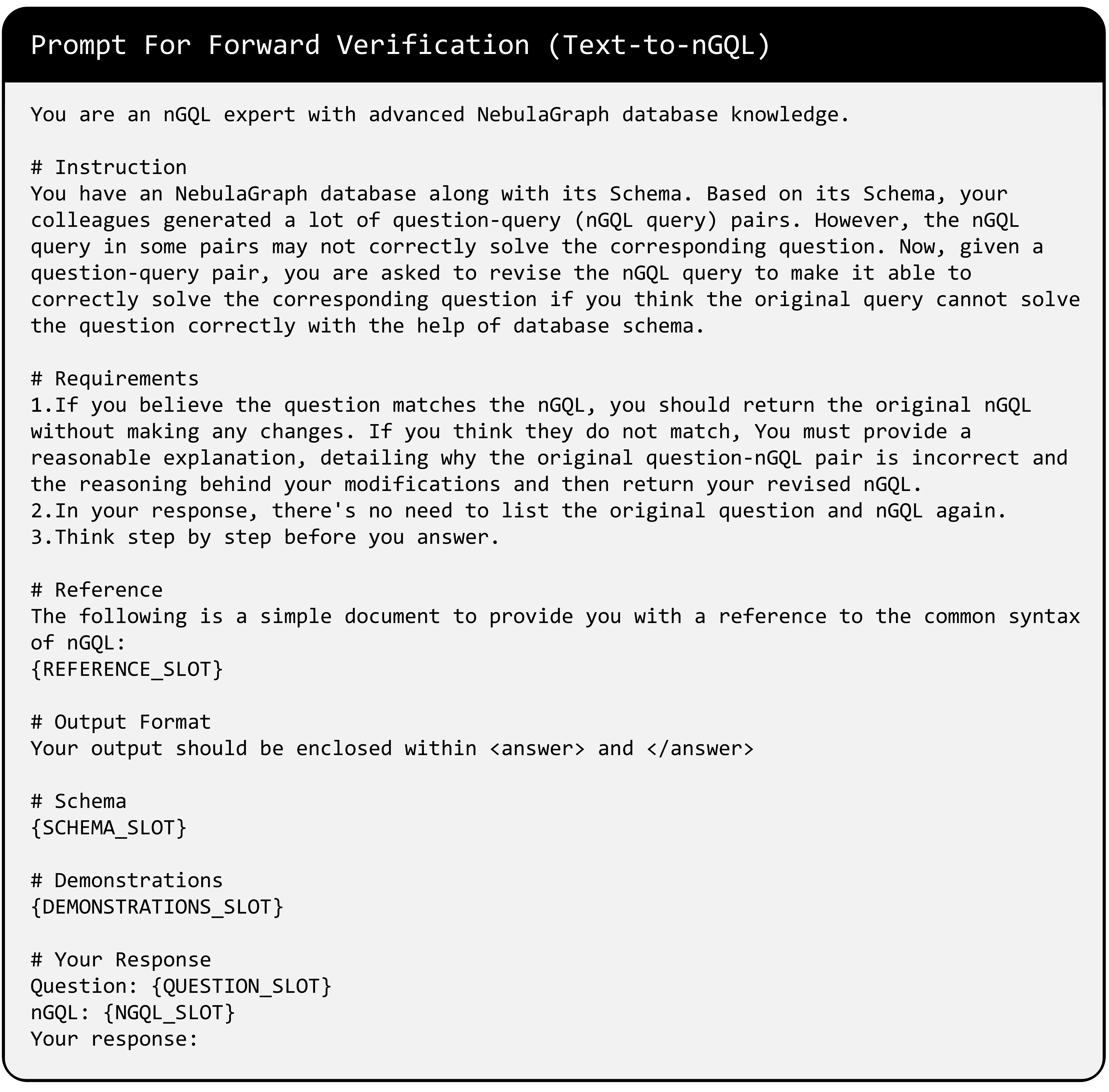}
\caption{Prompt for Forward Verification in the Data Synthesis Pipeline of Text-to-nGQL.}
\end{figure*}
\end{document}